# Velocity/Position Integration Formula (I): Application to In-flight Coarse Alignment

Yuanxin Wu and Xianfei Pan

**Abstract—The in-flight alignment is a critical stage for airborne INS/GPS applications. The alignment task is usually carried out by the Kalman filtering technique that necessitates a good initial attitude to obtain satisfying performance. Due to the airborne dynamics, the in-flight alignment is much difficult than alignment on the ground. This paper proposes an optimization-based coarse alignment approach using GPS position/velocity as input, founded on the newly-derived velocity/position integration formulae. Simulation and flight test results show that, with the GPS lever arm well handled, it is potentially able to yield the initial heading up to one degree accuracy in ten seconds. It can serve as a nice coarse in-flight alignment without any prior attitude information for the subsequent fine Kalman alignment. The approach can also be applied to other applications that require aligning the INS on the run.**

## I. INTRODUCTION

Initial alignment is vitally important because the performance of a strapdown inertial navigation system (SINS) is largely decided by the accuracy and rapidness of the alignment process. Since other initial conditions such as position and velocity are relatively easy to acquire, e.g., by GPS, we care most about the SINS body attitude with respect to the reference navigation frame during the alignment stage.

The alignment procedure often consists of two consecutive stages: coarse alignment and fine alignment. The current fine alignment methods based on Kalman filtering rely heavily on the coarse alignment stage to provide a

This work was supported in part by the Fok Ying Tung Foundation (131061), National Natural Science Foundation of China (61174002), the Foundation for the Author of National Excellent Doctoral Dissertation of People's Republic of China (FANEDD 200897) and Program for New Century Excellent Talents in University (NCET-10-0900). A short version was presented at IGNSS' 2011, Sydney, Australia.

Authors' address: Department of Automatic Control, College of Mechatronics and Automation, National University of Defense Technology, Changsha, Hunan, P. R. China, 410073. Tel/Fax: 086-0731-84576305-8212 (e-mail: yuanx_wu@hotmail.com).

roughly known initial attitude, otherwise they cannot guarantee a rapid and accurate alignment result [1-5]. If the SINS is stationary or quasi-stationary, analytic methods are often used to derive a coarse attitude from gyroscope/accelerometer measurements [1, 3, 6]. The heading angle is more difficult to determine than the two level angles and for a consumer grade SINS, is usually aided by a magnetic compass. The in-motion or in-flight alignment is necessary for many military applications and commercial aviations [7]. In such cases, the SINS is in motion, e.g., onboard a ship or aircraft, the direction of velocity/trajectory from an aided source, such as GPS, can provide a rough pitch and heading angles during a straight course. This information is generally not good enough to perform a reliable fine alignment due to the water current and air speed [3], let alone the SINS misaligning angles relative to the carrier.

It may be argued that the coarse alignment difficulty confronting the navigation field is largely owed to our "local eye" on the attitude representation in three-dimension space. Nowadays, we are used to the attitude approximation by three one-dimension error angles. For example, many works have been devoted to the nonlinear angle error models to account for large heading uncertainty [4, 8, 9]. By so doing, most inherent characteristics of the three-dimension attitude have been lost. Our group proposed a recursive alignment approach based on attitude optimization in [10], which, for the first time in the public literature, transforms the attitude alignment problem into a continuous attitude determination problem [11] using infinite vector observations. It was rigorously proven therein that the behavior of the estimated constant initial angles can be used to detect significant sensor biases. The optimization approach is related to the so-called inertial frame method [12, 13], but more theoretically solid and more robust to disturbances and noise, because it makes full use of the special algebraic property of the attitude matrix (a three-dimension orthogonal matrix with unit determinant). If the nonzero velocity rate information was externally provided, the optimization approach [10] could be readily applied to the in-flight alignment. The recent work [14] investigated the in-flight alignment in a quite similar manner based on the interleaved GPS-sample time integral of measured motions. In [15], we have independently devised the so-called velocity integration formula to circumvent the velocity rate calculation, necessary in [10], from ubiquitously noisy source information and reported preliminary flight test results.

This paper systematically and comprehensively extends the research in [15] by the devised the velocity/position integration formulae. The companion paper [16] applies the incremental forms of the two integration formulae to strapdown inertial navigation computation. Recursive in-flight alignment algorithms are delicately designed and evaluated using numerical and flight test data. The contents are organized as follows. Section II mathematically makes the alignment problem statement. Section III devises the velocity/position integration formulae and the recursive discrete algorithms respectively based on the two integration formulae are designed in Section IV. Simulations and real flight test results are reported in Section V and Section VI. Finally, conclusions are drawn in Section VII. The contribution of this paper, along with [10, 15], is proposing a systematic approach to solve the SINS alignment in motion using absolute velocity/position sensors.

## II. Problem Statement

Denote by $N$ the local level navigation frame, by $B$ the SINS body frame, by $I$ the inertially non-rotating frame, by $E$ the Earth frame. The navigation (attitude, velocity and position) rate equations in the $N$-frame are respectively known as [1, 3, 17]

$$\dot{\mathbf{C}}_b^n = \mathbf{C}_b^n \boldsymbol{\omega}_{nb}^b \times \tag{1}$$

$$\dot{\mathbf{v}}^n = \mathbf{C}_b^n \mathbf{f}^b - \left(2\boldsymbol{\omega}_{ie}^n + \boldsymbol{\omega}_{en}^n\right) \times \mathbf{v}^n + \mathbf{g}^n \tag{2}$$

$$\dot{\mathbf{p}} = \mathbf{R}_c \mathbf{v}^n \tag{3}$$

where $\mathbf{C}_b^n$ denotes the attitude matrix from the body frame to the navigation frame, $\mathbf{v}^n$ the velocity relative to the Earth (also called ground velocity), $\boldsymbol{\omega}_{ib}^b$ the body angular rate measured by gyroscopes in the body frame, $\mathbf{f}^b$ the specific force measured by accelerometers in the body frame, $\boldsymbol{\omega}_{ie}^n$ the Earth rotation rate with respect to the inertial frame, $\boldsymbol{\omega}_{en}^n$ the angular rate of the navigation frame with respect to the Earth frame, $\boldsymbol{\omega}_{nb}^b = \boldsymbol{\omega}_{ib}^b - \mathbf{C}_n^b \boldsymbol{\omega}_{in}^n$ is the body angular rate with respect to the navigation frame, and $\mathbf{g}^n$ is the gravity vector. The $3 \times 3$ skew symmetric matrix $(\cdot \times)$ is defined so that the cross product satisfies $\mathbf{a} \times \mathbf{b} = (\mathbf{a} \times)\mathbf{b}$ for arbitrary two vectors. The position $\mathbf{p} \triangleq \begin{bmatrix} \lambda & L & h \end{bmatrix}^T$ is described by the angular orientation of the navigation frame relative to the Earth frame, commonly expressed as longitude $\lambda$ and latitude $L$, and the height above the Earth surface $h$. $\mathbf{R}_c$ is the local curvature matrix that is a function of the current position and defined explicitly in (13). All the quantities above are functions of time and, if not stated, their time dependences on are omitted for brevity.

Suppose the alignment process starts from $t = 0$. Given the measured velocity $\mathbf{v}^n$ and position $\mathbf{p}$ over the time interval of interest $\left[0,\ t\right]$, our purpose is to determine the attitude $\mathbf{C}_b^n$ while in motion.

## III. VELOCITY/POSITION INTEGRATION FORMULAE

### A. Attitude Integration Formula

Equation (1) shows that the attitude matrix $\mathbf{C}_b^n$ is a function of $\boldsymbol{\omega}_{nb}^b$, whose calculation inversely depends on $\mathbf{C}_b^n$ by the relation $\boldsymbol{\omega}_{nb}^b = \boldsymbol{\omega}_{ib}^b - \mathbf{C}_n^b\boldsymbol{\omega}_{in}^n$. In order to obtain the analytic form of $\mathbf{C}_b^n$, we separately consider the attitude changes of the body frame and the navigation frame relative to some chosen inertial frame and then combine them together [10]. Specifically, by the chain rule of the attitude matrix, $\mathbf{C}_b^n$ at any time satisfies

$$\mathbf{C}_b^n\left(t\right) = \mathbf{C}_{b(t)}^{n(t)} = \mathbf{C}_{n(0)}^{n(t)}\mathbf{C}_{b(0)}^{n(0)}\mathbf{C}_{b(t)}^{b(0)} = \mathbf{C}_{n(0)}^{n(t)}\mathbf{C}_b^n\left(0\right)\mathbf{C}_{b(t)}^{b(0)} \tag{4}$$

where $\mathbf{C}_{b(0)}^{b(t)}$ and $\mathbf{C}_{n(0)}^{n(t)}$ respectively encode the attitude changes of the body frame and the navigation frame from time $0$ to $t$. Their rate equations are

$$\begin{aligned}
\dot{\mathbf{C}}_{b(t)}^{b(0)} &= \mathbf{C}_{b(t)}^{b(0)}\boldsymbol{\omega}_{ib}^b \times \\
\dot{\mathbf{C}}_{n(t)}^{n(0)} &= \mathbf{C}_{n(t)}^{n(0)}\boldsymbol{\omega}_{in}^n \times
\end{aligned} \tag{5}$$

If the initial attitude matrix $\mathbf{C}_b^n\left(0\right)$, which is a constant quantity, was known, then $\mathbf{C}_b^n$ at any time can be obtained by (4) and (5).

Note that for a fixed time $t_0$, both the body frame and the navigation frame with respect to any $I$-frame, say $\mathbf{C}_i^{b(t_0)}$ and $\mathbf{C}_i^{n(t_0)}$, are functions of $t_0$ instead of $t$, and hence their time derivatives are zero. To put it the other way, they are inertially "frozen" after the time epoch $t_0$ passes. Compared to the attitude differential equation in (1), (4) presents the explicit form of the attitude matrix $\mathbf{C}_b^n$ and is thus referred to as the attitude integration formula [10]. Its incremental version over one update interval is not unfamiliar in the previous literature, see e.g., [18].

### B. Velocity Integration Formula

The velocity integration formula can be achieved by basic integration transformations. Substituting (4) into (2) yields

$$\dot{\mathbf{v}}^n = \mathbf{C}_{n(0)}^{n(t)} \mathbf{C}_b^n (0) \mathbf{C}_{b(t)}^{b(0)} \mathbf{f}^b - \left(2\boldsymbol{\omega}_{ie}^n + \boldsymbol{\omega}_{en}^n\right) \times \mathbf{v}^n + \mathbf{g}^n \tag{6}$$

Multiplying $\mathbf{C}_{n(t)}^{n(0)}$ on both sides,

$$\mathbf{C}_{n(t)}^{n(0)} \dot{\mathbf{v}}^n = \mathbf{C}_b^n (0) \mathbf{C}_{b(t)}^{b(0)} \mathbf{f}^b - \mathbf{C}_{n(t)}^{n(0)} \left(2\boldsymbol{\omega}_{ie}^n + \boldsymbol{\omega}_{en}^n\right) \times \mathbf{v}^n + \mathbf{C}_{n(t)}^{n(0)} \mathbf{g}^n \tag{7}$$

or equivalently,

$$\mathbf{C}_b^n (0) \mathbf{C}_{b(t)}^{b(0)} \mathbf{f}^b = \mathbf{C}_{n(t)}^{n(0)} \left(\dot{\mathbf{v}}^n + \left(2\boldsymbol{\omega}_{ie}^n + \boldsymbol{\omega}_{en}^n\right) \times \mathbf{v}^n - \mathbf{g}^n\right) \tag{8}$$

Equation (8) forms the basis for the optimization approach [10], which could be readily applied to the in-flight alignment problem discussed in this paper if the velocity rate $\dot{\mathbf{v}}^n$ was well provided in addition to velocity and position information.

Integrating (7) on both sides over the time interval of interest,

$$\int_0^t \mathbf{C}_{n(t)}^{n(0)} \dot{\mathbf{v}}^n dt = \mathbf{C}_b^n (0) \int_0^t \mathbf{C}_{b(t)}^{b(0)} \mathbf{f}^b dt - \int_0^t \mathbf{C}_{n(t)}^{n(0)} \left(2\boldsymbol{\omega}_{ie}^n + \boldsymbol{\omega}_{en}^n\right) \times \mathbf{v}^n dt + \int_0^t \mathbf{C}_{n(t)}^{n(0)} \mathbf{g}^n dt \tag{9}$$

The left term is developed as

$$\int_0^t \mathbf{C}_{n(t)}^{n(0)} \dot{\mathbf{v}}^n dt = \mathbf{C}_{n(t)}^{n(0)} \mathbf{v}^n \Big|_0^t - \int_0^t \mathbf{C}_{n(t)}^{n(0)} \boldsymbol{\omega}_{in}^n \times \mathbf{v}^n dt = \mathbf{C}_{n(t)}^{n(0)} \mathbf{v}^n - \mathbf{v}^n (0) - \int_0^t \mathbf{C}_{n(t)}^{n(0)} \boldsymbol{\omega}_{in}^n \times \mathbf{v}^n dt \tag{10}$$

where the attitude rate equation (5) is used and $\mathbf{v}^n (0)$ is the initial velocity. Substituting (10) into (9) and reorganizing the terms yield the velocity integration formula as

$$\mathbf{C}_b^n (0) \boldsymbol{\alpha}_v = \boldsymbol{\beta}_v \tag{11}$$

in which

$$\begin{aligned}
\boldsymbol{\alpha}_v &\triangleq \int_0^t \mathbf{C}_{b(0)}^{b(0)} \mathbf{f}^b dt \\
\boldsymbol{\beta}_v &\triangleq \mathbf{C}_{n(t)}^{n(0)} \mathbf{v}^n - \mathbf{v}^n (0) + \int_0^t \mathbf{C}_{n(t)}^{n(0)} \boldsymbol{\omega}_{ie}^n \times \mathbf{v}^n dt - \int_0^t \mathbf{C}_{n(t)}^{n(0)} \mathbf{g}^n dt
\end{aligned} \tag{12}$$

Equation (11) is an integration function of the initial attitude matrix $\mathbf{C}_b^n (0)$ for any $t$ and the two vectors, $\boldsymbol{\alpha}_v$ and $\boldsymbol{\beta}_v$, are evidently defined. It can be shown that (11) is the integral form of the velocity rate equation in the inertial frame mechanism (p. 28, [1]) that was directly used by [14] in its incremental form over interleaved one-second integral intervals. Note that $\boldsymbol{\alpha}_v$ and $\boldsymbol{\beta}_v$ are respectively functions of the gyroscope/accelerometer outputs and the aided velocity/position information during the alignment period $\begin{bmatrix} 0, \ t \end{bmatrix}$. Theoretically, if there exist two linearly independent $\boldsymbol{\alpha}_v$ (or $\boldsymbol{\beta}_v$), the initial attitude matrix $\mathbf{C}_b^n(0)$ can be uniquely determined [11]. This condition is easily fulfilled in practice and readers are referred to [10] for more analysis.

## C. Position Integration Formula

In the context of a specific local level frame choice, e.g., North-Up-East, $\mathbf{v}^n = \begin{bmatrix} v_N & v_U & v_E \end{bmatrix}^T$, the local curvature matrix is explicitly expressed as a function of current position

$$\mathbf{R}_c = \begin{bmatrix} 0 & 0 & \dfrac{1}{(R_E+h)\cos L} \\ \dfrac{1}{R_N+h} & 0 & 0 \\ 0 & 1 & 0 \end{bmatrix} \text{ or } \mathbf{R}_c^{-1} = \begin{bmatrix} 0 & R_N+h & 0 \\ 0 & 0 & 1 \\ (R_E+h)\cos L & 0 & 0 \end{bmatrix} \quad (13)$$

where $R_E$ and $R_N$ are respectively the transverse radius of curvature and the meridian radius of curvature of the WGS-84 reference ellipsoid. The specific expression of $\mathbf{R}_c$ will be different for other local level frame choices but it does not hinder from understanding the main idea of this paper.

Rewrite (3) as

$$\mathbf{R}_c^{-1}\dot{\mathbf{p}} = \mathbf{v}^n \quad (14)$$

Define the position in *N*-frame as the integral

$$\mathbf{r}^n(t) = \int_0^t \mathbf{v}^n dt = \int_0^t \mathbf{R}_c^{-1}\dot{\mathbf{p}} dt \quad (15)$$

whose rate equation is given as

$$\dot{\mathbf{r}}^n = \mathbf{v}^n \quad (16)$$

Note that $\mathbf{r}^n(0) = 0$.

Regarding (11), we have from (16)

$$\mathbf{C}_{n(t)}^{n(0)}\dot{\mathbf{r}}^n = \mathbf{C}_{n(t)}^{n(0)}\mathbf{v}^n = \mathbf{v}^n(0) + \mathbf{C}_b^n(0)\int_0^t \mathbf{C}_{b(t)}^{b(0)}\mathbf{f}^b dt - \int_0^t \mathbf{C}_{n(t)}^{n(0)}\boldsymbol{\omega}_{ie}^n \times \mathbf{v}^n dt + \int_0^t \mathbf{C}_{n(t)}^{n(0)}\mathbf{g}^n dt \quad (17)$$

By the same techniques as in (9)

$$\int_0^t \mathbf{C}_{n(t)}^{n(0)} \dot{\mathbf{r}}^n dt = \mathbf{C}_{n(t)}^{n(0)} \mathbf{r}^n \Big|_0^t - \int_0^t \mathbf{C}_{n(t)}^{n(0)} \boldsymbol{\omega}_{in}^n \times \mathbf{r}^n dt = \mathbf{C}_{n(t)}^{n(0)} \mathbf{r}^n - \int_0^t \mathbf{C}_{n(t)}^{n(0)} \boldsymbol{\omega}_{in}^n \times \mathbf{r}^n dt \tag{18}$$

Integrating (17) over the time interval $\begin{bmatrix} 0, & t \end{bmatrix}$, we obtain

$$\int_0^t \mathbf{C}_{n(t)}^{n(0)} \mathbf{v}^n dt = t\,\mathbf{v}^n(0) + \mathbf{C}_b^n(0)\int_0^t \int_0^\tau \mathbf{C}_{b(\sigma)}^{b(0)}\,\mathbf{f}^b d\sigma d\tau - \int_0^t \int_0^\tau \mathbf{C}_{n(\sigma)}^{n(0)} \boldsymbol{\omega}_{ie}^n \times \mathbf{v}^n d\sigma d\tau + \int_0^t \int_0^\tau \mathbf{C}_{n(\sigma)}^{n(0)} \mathbf{g}^n d\sigma d\tau \tag{19}$$

Organizing the terms yield the position integration formula as

$$\mathbf{C}_b^n(0)\boldsymbol{\alpha}_p = \boldsymbol{\beta}_p \tag{20}$$

in which

$$\begin{aligned}
\boldsymbol{\alpha}_p &\triangleq \int_0^t \int_0^\tau \mathbf{C}_{b(\sigma)}^{b(0)}\,\mathbf{f}^b d\sigma d\tau \\
\boldsymbol{\beta}_p &\triangleq \int_0^t \mathbf{C}_{n(t)}^{n(0)} \mathbf{v}^n dt - t\,\mathbf{v}^n(0) + \int_0^t \int_0^\tau \mathbf{C}_{n(\sigma)}^{n(0)} \boldsymbol{\omega}_{ie}^n \times \mathbf{v}^n d\sigma d\tau - \int_0^t \int_0^\tau \mathbf{C}_{n(\sigma)}^{n(0)} \mathbf{g}^n d\sigma d\tau
\end{aligned} \tag{21}$$

Dividing the nonzero time length on both sides,

$$\begin{aligned}
\mathbf{C}_b^n(0)\frac{\boldsymbol{\alpha}_p}{t} &= \mathbf{C}_b^n(0)\frac{1}{t}\int_0^t \int_0^\tau \mathbf{C}_{b(\sigma)}^{b(0)}\,\mathbf{f}^b d\sigma d\tau \\
&= \frac{1}{t}\left(\int_0^t \mathbf{C}_{n(t)}^{n(0)} \mathbf{v}^n dt + \int_0^t \int_0^\tau \mathbf{C}_{n(\sigma)}^{n(0)} \boldsymbol{\omega}_{ie}^n \times \mathbf{v}^n d\sigma d\tau - \int_0^t \int_0^\tau \mathbf{C}_{n(\sigma)}^{n(0)} \mathbf{g}^n d\sigma d\tau\right) - \mathbf{v}^n(0) = \frac{\boldsymbol{\beta}_p}{t}
\end{aligned} \tag{22}$$

where $\mathbf{v}^n(0)$ appears as an additive term, the same as in (11) and the vectors $\boldsymbol{\alpha}_p$ and $\boldsymbol{\beta}_p$ are defined obviously.

Equations (20)-(22) are also an integration functions of the initial attitude matrix $\mathbf{C}_b^n(0)$ for any positive $t$, in which $\boldsymbol{\alpha}_p$ and $\boldsymbol{\beta}_p$ are respectively functions of the gyroscope/accelerometer outputs and the aided velocity/position information during the alignment period $\begin{bmatrix} 0, & t \end{bmatrix}$.

*Remark*: Equation (22) is not simply an integration of (11) in that it has a simplified form that does not need the aided velocity as an input. Substituting (18) into (22) gives

$$\begin{aligned}
\mathbf{C}_b^n(0)\frac{\boldsymbol{\alpha}_p}{t} &= \mathbf{C}_b^n(0)\frac{1}{t}\int_0^t \int_0^\tau \mathbf{C}_{b(\sigma)}^{b(0)}\,\mathbf{f}^b d\sigma d\tau \\
&= \frac{1}{t}\left(\mathbf{C}_{n(t)}^{n(0)} \mathbf{r}^n - \int_0^t \mathbf{C}_{n(t)}^{n(0)} \boldsymbol{\omega}_{in}^n \times \mathbf{r}^n dt + \int_0^t \int_0^\tau \mathbf{C}_{n(\sigma)}^{n(0)} \boldsymbol{\omega}_{ie}^n \times \mathbf{v}^n d\sigma d\tau - \int_0^t \int_0^\tau \mathbf{C}_{n(\sigma)}^{n(0)} \mathbf{g}^n d\sigma d\tau\right) - \mathbf{v}^n(0) = \frac{\boldsymbol{\beta}_p}{t}
\end{aligned} \tag{23}$$

which have two velocity-related terms, namely, the double integral $\int_0^t \int_0^\tau \mathbf{C}_{n(\sigma)}^{n(0)} \boldsymbol{\omega}_{ie}^n \times \mathbf{v}^n d\sigma d\tau$ and the additive initial velocity $\mathbf{v}^n(0)$. The former is relatively small and can be omitted, and the latter can be eliminated by the difference between any time instants. In addition, $\boldsymbol{\omega}_{in}^n$ can be well approximated by $\boldsymbol{\omega}_{ie}^n$ for most applications. These approximations may allow for a coarse in-flight alignment using position information only.

## IV. RECURSIVE IN-FLIGHT ALGORITHMS

In this section, we will design delicate numerical algorithms so as to accurately calculate the integrals involved in (11) and (20) and then solve the alignment problem using attitude optimization [10]. Suppose the current time is $M$, an integer, times of the updated interval, i.e., $t \triangleq MT$, where $T$ is the time duration of the update interval $[t_k \quad t_{k+1}]$, $k = 0, 1, 2, ..., M-1$ and $t_k = kT$.

### A. Calculating Integrals in Velocity Integration Formula

The last integral in (11) can be written as

$$\int_0^t \mathbf{C}_{n(t)}^{n(0)} \mathbf{g}^n dt = \sum_{k=0}^{M-1} \int_{t_k}^{t_{k+1}} \mathbf{C}_{n(t)}^{n(0)} \mathbf{g}^n dt = \sum_{k=0}^{M-1} \mathbf{C}_{n(t_k)}^{n(0)} \int_{t_k}^{t_{k+1}} \mathbf{C}_{n(t)}^{n(t_k)} \mathbf{g}^n dt \tag{24}$$

Because $\boldsymbol{\omega}_{in}^n$ is usually slowly changing, we approximate the attitude matrix by

$\mathbf{C}_{n(t)}^{n(t_k)} = I + \dfrac{\sin(\|\boldsymbol{\varphi}_n\|)}{\|\boldsymbol{\varphi}_n\|} \boldsymbol{\varphi}_n \times + \dfrac{1 - \cos(\|\boldsymbol{\varphi}_n\|)}{\|\boldsymbol{\varphi}_n\|^2} (\boldsymbol{\varphi}_n \times)^2 \approx \mathbf{I} + \boldsymbol{\varphi}_n \times$ , where $\boldsymbol{\varphi}_n \approx \int_{t_k}^t \boldsymbol{\omega}_{in}^n d\tau \approx (t - t_k) \boldsymbol{\omega}_{in}^n$ denotes the

$N$-frame rotation vector from $t_k$ to the current time. So, the last integral is approximated by

$$\int_0^t \mathbf{C}_{n(t)}^{n(0)} \mathbf{g}^n dt \approx \sum_{k=0}^{M-1} \mathbf{C}_{n(t_k)}^{n(0)} \int_{t_k}^{t_{k+1}} \left(\mathbf{I} + (t - t_k) \boldsymbol{\omega}_{in}^n \times\right) \mathbf{g}^n dt = \sum_{k=0}^{M-1} \mathbf{C}_{n(t_k)}^{n(0)} \left(T\mathbf{I} + \dfrac{T^2}{2} \boldsymbol{\omega}_{in}^n \times\right) \mathbf{g}^n \tag{25}$$

where the quantities $\boldsymbol{\omega}_{in}^n$ and $\mathbf{g}^n$ can be approximately regarded as constants during the incremental interval and take the values at the lower integral limit $t_k$.

Suppose the velocity $\mathbf{v}^n$ changes linearly during $[t_k, \quad t_{k+1}]$, i.e.,

$$\mathbf{v}^n(t) = \mathbf{v}^n(t_k) + \dfrac{t - t_k}{T} \left(\mathbf{v}^n(t_{k+1}) - \mathbf{v}^n(t_k)\right) \tag{26}$$

The first integral on the right side of (11) is approximated by

$$\int_0^t \mathbf{C}_{n(t)}^{n(0)} \boldsymbol{\omega}_{ie}^n \times \mathbf{v}^n dt = \sum_{k=0}^{M-1} \mathbf{C}_{n(t)}^{n(0)} \int_{t_k}^{t_{k+1}} \mathbf{C}_{n(t)}^{n(t_k)} \boldsymbol{\omega}_{ie}^n \times \mathbf{v}^n dt$$

$$\approx \sum_{k=0}^{M-1} \mathbf{C}_{n(t_k)}^{n(0)} \int_{t_k}^{t_{k+1}} \left( \mathbf{I} + (t - t_k)\, \boldsymbol{\omega}_{in}^n \times \right) \boldsymbol{\omega}_{ie}^n \times \left( \mathbf{v}^n(t_k) + \frac{t - t_k}{T} \left( \mathbf{v}^n(t_{k+1}) - \mathbf{v}^n(t_k) \right) \right) dt$$

$$= \sum_{k=0}^{M-1} \mathbf{C}_{n(t_k)}^{n(0)} \left[ \left( T\mathbf{I} + \frac{T^2}{2} \boldsymbol{\omega}_{in}^n \times \right) \boldsymbol{\omega}_{ie}^n \times \mathbf{v}^n(t_k) + \left( \frac{T}{2}\mathbf{I} + \frac{T^2}{3} \boldsymbol{\omega}_{in}^n \times \right) \boldsymbol{\omega}_{ie}^n \times \left( \mathbf{v}^n(t_{k+1}) - \mathbf{v}^n(t_k) \right) \right]$$

$$= \sum_{k=0}^{M-1} \mathbf{C}_{n(t_k)}^{n(0)} \left[ \left( \frac{T}{2}\mathbf{I} + \frac{T^2}{6} \boldsymbol{\omega}_{in}^n \times \right) \boldsymbol{\omega}_{ie}^n \times \mathbf{v}^n(t_k) + \left( \frac{T}{2}\mathbf{I} + \frac{T^2}{3} \boldsymbol{\omega}_{in}^n \times \right) \boldsymbol{\omega}_{ie}^n \times \mathbf{v}^n(t_{k+1}) \right]$$

(27)

where the quantities $\boldsymbol{\omega}_{ie}^n$ is also regarded as constants during the interval of interest.

Substituting (25) and (27) into the right side of (11),

$$\boldsymbol{\beta}_v(t_M) = \mathbf{C}_{n(t_M)}^{n(0)} \mathbf{v}^n - \mathbf{v}^n(0)$$

$$+ \sum_{k=0}^{M-1} \mathbf{C}_{n(t_k)}^{n(0)} \left[ \left( \frac{T}{2}\mathbf{I} + \frac{T^2}{6} \boldsymbol{\omega}_{in}^n \times \right) \boldsymbol{\omega}_{ie}^n \times \mathbf{v}^n(t_k) + \left( \frac{T}{2}\mathbf{I} + \frac{T^2}{3} \boldsymbol{\omega}_{in}^n \times \right) \boldsymbol{\omega}_{ie}^n \times \mathbf{v}^n(t_{k+1}) - \left( T\mathbf{I} + \frac{T^2}{2} \boldsymbol{\omega}_{in}^n \times \right) \mathbf{g}^n \right]$$

(28)

The vector $\boldsymbol{\alpha}_v$ in (11) can be approximated as

$$\boldsymbol{\alpha}_v(t_M) = \int_0^t \mathbf{C}_{b(t)}^{b(0)} \mathbf{f}^b dt$$

$$= \sum_{k=0}^{M-1} \mathbf{C}_{b(t_k)}^{b(0)} \int_{t_k}^{t_{k+1}} \mathbf{C}_{b(t)}^{b(t_k)} \mathbf{f}^b dt \approx \sum_{k=0}^{M-1} \mathbf{C}_{b(t_k)}^{b(0)} \int_{t_k}^{t_{k+1}} \left( \mathbf{I} + \left( \int_{t_k}^t \boldsymbol{\omega}_{ib}^b d\tau \right) \times \right) \mathbf{f}^b dt$$

(29)

where the incremental integral above can be approximated using the two-sample correction by (see Appendix A)

$$\int_{t_k}^{t_{k+1}} \left( \mathbf{I} + \left( \int_{t_k}^t \boldsymbol{\omega}_{ib}^b d\tau \right) \times \right) \mathbf{f}^b dt = \Delta \mathbf{v}_1 + \Delta \mathbf{v}_2 + \frac{1}{2} \left( \Delta \boldsymbol{\theta}_1 + \Delta \boldsymbol{\theta}_2 \right) \times \left( \Delta \mathbf{v}_1 + \Delta \mathbf{v}_2 \right) + \frac{2}{3} \left( \Delta \boldsymbol{\theta}_1 \times \Delta \mathbf{v}_2 + \Delta \mathbf{v}_1 \times \Delta \boldsymbol{\theta}_2 \right)$$

(30)

where $\Delta \mathbf{v}_1, \Delta \mathbf{v}_2$ are the first and second samples of the accelerometer-measured incremental velocity and $\Delta \boldsymbol{\theta}_1, \Delta \boldsymbol{\theta}_2$ are the first and second samples of the gyroscope-measured incremental angle, respectively. In contrast to $\boldsymbol{\omega}_{in}^n$, the body rate $\boldsymbol{\omega}_{ib}^b$ is fast changing and has to be handled by special treatment as such. The technique is a common practice in the inertial navigation community [1, 3, 17].

### B. Calculating Integrals in Position Integration Formula

Next, we will handle the integrals involved in the position integration formula. With (25), the second double integral on the right side of (20) can be approximated by

$$\int_0^t \int_0^\tau \mathbf{C}_{n(\sigma)}^{n(0)} \mathbf{g}^n d\sigma d\tau = \sum_{k=0}^{M-1} \int_{t_k}^{t_{k+1}} \int_0^\tau \mathbf{C}_{n(\sigma)}^{n(0)} \mathbf{g}^n d\sigma d\tau = \sum_{k=0}^{M-1} \int_{t_k}^{t_{k+1}} \left( \int_0^{t_k} \mathbf{C}_{n(\sigma)}^{n(0)} \mathbf{g}^n d\sigma + \int_{t_k}^\tau \mathbf{C}_{n(\sigma)}^{n(0)} \mathbf{g}^n d\sigma \right) d\tau$$

$$= \sum_{k=0}^{M-1} T \int_0^{t_k} \mathbf{C}_{n(\sigma)}^{n(0)} \mathbf{g}^n d\sigma + \int_{t_k}^{t_{k+1}} \int_{t_k}^\tau \mathbf{C}_{n(\sigma)}^{n(0)} \mathbf{g}^n d\sigma d\tau$$

$$= \sum_{k=0}^{M-1} \left( T \sum_{m=0}^{k-1} \mathbf{C}_{n(t_m)}^{n(0)} \int_{t_m}^{t_{m+1}} \mathbf{C}_{n(\sigma)}^{n(t_m)} \mathbf{g}^n d\sigma + \mathbf{C}_{n(t_k)}^{n(0)} \int_{t_k}^{t_{k+1}} \int_{t_k}^\tau \mathbf{C}_{n(\sigma)}^{n(t_k)} \mathbf{g}^n d\sigma d\tau \right) \tag{31}$$

$$= \sum_{k=0}^{M-1} \left( T \sum_{m=0}^{k-1} \mathbf{C}_{n(t_m)}^{n(0)} \int_{t_m}^{t_{m+1}} \left( \mathbf{I} + (\sigma - t_m) \boldsymbol{\omega}_{in}^n \times \right) \mathbf{g}^n d\sigma + \mathbf{C}_{n(t_k)}^{n(0)} \int_{t_k}^{t_{k+1}} \int_{t_k}^\tau \left( \mathbf{I} + (\sigma - t_k) \boldsymbol{\omega}_{in}^n \times \right) \mathbf{g}^n d\sigma d\tau \right)$$

$$= \sum_{k=0}^{M-1} \left( T \sum_{m=0}^{k-1} \mathbf{C}_{n(t_m)}^{n(0)} \left( T\mathbf{I} + \frac{T^2}{2} \boldsymbol{\omega}_{in}^n \times \right) \mathbf{g}^n + \mathbf{C}_{n(t_k)}^{n(0)} \left( \frac{T^2}{2} \mathbf{I} + \frac{T^3}{6} \boldsymbol{\omega}_{in}^n \times \right) \mathbf{g}^n \right)$$

Similarly with (26)-(27), the first double integral on the right side of (20) can be approximated by

$$\int_0^t \int_0^\tau \mathbf{C}_{n(\sigma)}^{n(0)} \boldsymbol{\omega}_{ie}^n \times \mathbf{v}^n d\sigma d\tau = \sum_{k=0}^{M-1} \left( T \sum_{m=0}^{k-1} \mathbf{C}_{n(t_m)}^{n(0)} \int_{t_m}^{t_{m+1}} \mathbf{C}_{n(\sigma)}^{n(t_m)} \boldsymbol{\omega}_{ie}^n \times \mathbf{v}^n d\sigma + \mathbf{C}_{n(t_k)}^{n(0)} \int_{t_k}^{t_{k+1}} \int_{t_k}^\tau \mathbf{C}_{n(\sigma)}^{n(t_k)} \boldsymbol{\omega}_{ie}^n \times \mathbf{v}^n d\sigma d\tau \right)$$

$$= \sum_{k=0}^{M-1} \left\{ T \sum_{m=0}^{k-1} \mathbf{C}_{n(t_m)}^{n(0)} \left[ \left( \frac{T}{2} \mathbf{I} + \frac{T^2}{6} \boldsymbol{\omega}_{in}^n \times \right) \boldsymbol{\omega}_{ie}^n \times \mathbf{v}^n (t_m) + \left( \frac{T}{2} \mathbf{I} + \frac{T^2}{3} \boldsymbol{\omega}_{in}^n \times \right) \boldsymbol{\omega}_{ie}^n \times \mathbf{v}^n (t_{m+1}) \right] \right. \tag{32}$$

$$\left. + \mathbf{C}_{n(t_k)}^{n(0)} \left[ \left( \frac{T^2}{3} \mathbf{I} + \frac{T^3}{12} \boldsymbol{\omega}_{in}^n \times \right) \boldsymbol{\omega}_{ie}^n \times \mathbf{v}^n (t_k) + \left( \frac{T^2}{6} \mathbf{I} + \frac{T^3}{12} \boldsymbol{\omega}_{in}^n \times \right) \boldsymbol{\omega}_{ie}^n \times \mathbf{v}^n (t_{k+1}) \right] \right\}$$

Assume linearly changing $\mathbf{v}^n$ as in (26), the single integral in (20) can be approximated by

$$\int_0^t \mathbf{C}_{n(\tau)}^{n(0)} \mathbf{v}^n d\tau = \sum_{k=0}^{M-1} \mathbf{C}_{n(t_k)}^{n(0)} \int_{t_k}^{t_{k+1}} \mathbf{C}_{n(\tau)}^{n(t_k)} \mathbf{v}^n d\tau$$

$$= \sum_{k=0}^{M-1} \mathbf{C}_{n(t_k)}^{n(0)} \left[ \left( \frac{T}{2} I + \frac{T^2}{6} \boldsymbol{\omega}_{in}^n \times \right) \mathbf{v}^n (t_k) + \left( \frac{T}{2} I + \frac{T^2}{3} \boldsymbol{\omega}_{in}^n \times \right) \mathbf{v}^n (t_{k+1}) \right] \tag{33}$$

Substituting (31)-(33) into the right side of (20), we have

$$\boldsymbol{\beta}_p (t_M) = \sum_{k=0}^{M-1} \mathbf{C}_{n(t_k)}^{n(0)} \left[ \left( \frac{T}{2} I + \frac{T^2}{6} \boldsymbol{\omega}_{in}^n \times \right) \mathbf{v}^n (t_k) + \left( \frac{T}{2} I + \frac{T^2}{3} \boldsymbol{\omega}_{in}^n \times \right) \mathbf{v}^n (t_{k+1}) \right] - t_M \mathbf{v}^n (0)$$

$$+ \sum_{k=0}^{M-1} \left\{ T \sum_{m=0}^{k-1} \mathbf{C}_{n(t_m)}^{n(0)} \left[ \left( \frac{T}{2} \mathbf{I} + \frac{T^2}{6} \boldsymbol{\omega}_{in}^n \times \right) \boldsymbol{\omega}_{ie}^n \times \mathbf{v}^n (t_m) + \left( \frac{T}{2} \mathbf{I} + \frac{T^2}{3} \boldsymbol{\omega}_{in}^n \times \right) \boldsymbol{\omega}_{ie}^n \times \mathbf{v}^n (t_{m+1}) \right] \right.$$

$$+ \mathbf{C}_{n(t_k)}^{n(0)} \left[ \left( \frac{T^2}{3} \mathbf{I} + \frac{T^3}{12} \boldsymbol{\omega}_{in}^n \times \right) \boldsymbol{\omega}_{ie}^n \times \mathbf{v}^n (t_k) + \left( \frac{T^2}{6} \mathbf{I} + \frac{T^3}{12} \boldsymbol{\omega}_{in}^n \times \right) \boldsymbol{\omega}_{ie}^n \times \mathbf{v}^n (t_{k+1}) \right] \right\} \tag{34}$$

$$- \sum_{k=0}^{M-1} \left( T \sum_{m=0}^{k-1} \mathbf{C}_{n(t_m)}^{n(0)} \left( T\mathbf{I} + \frac{T^2}{2} \boldsymbol{\omega}_{in}^n \times \right) \mathbf{g}^n + \mathbf{C}_{n(t_k)}^{n(0)} \left( \frac{T^2}{2} \mathbf{I} + \frac{T^3}{6} \boldsymbol{\omega}_{in}^n \times \right) \mathbf{g}^n \right)$$

The left side of (20) is approximated by

$$\boldsymbol{\alpha}_p\left(t_M\right)=\int_0^t\int_0^\tau \mathbf{C}_{b(\sigma)}^{b(0)}\,\mathbf{f}^b d\sigma d\tau=\sum_{k=0}^{M-1}\int_{t_k}^{t_{k+1}}\int_0^\tau \mathbf{C}_{b(\sigma)}^{b(0)}\,\mathbf{f}^b d\sigma d\tau$$

$$=\sum_{k=0}^{M-1}\int_{t_k}^{t_{k+1}}\left(\int_0^{t_k}\mathbf{C}_{b(\sigma)}^{b(0)}\,\mathbf{f}^b d\sigma+\int_{t_k}^\tau \mathbf{C}_{b(\sigma)}^{b(0)}\,\mathbf{f}^b d\sigma\right)d\tau$$

$$=\sum_{k=0}^{M-1}\left[T\int_0^{t_k}\mathbf{C}_{b(\sigma)}^{b(0)}\,\mathbf{f}^b d\sigma+\int_{t_k}^{t_{k+1}}\int_{t_k}^\tau \mathbf{C}_{b(\sigma)}^{b(0)}\,\mathbf{f}^b d\sigma d\tau\right]$$

$$=\sum_{k=0}^{M-1}\left[T\sum_{m=0}^{k-1}\mathbf{C}_{b(t_m)}^{b(0)}\int_{t_m}^{t_{m+1}}\mathbf{C}_{b(\sigma)}^{b(t_m)}\,\mathbf{f}^b d\sigma+\mathbf{C}_{b(t_k)}^{b(0)}\int_{t_k}^{t_{k+1}}\int_{t_k}^\tau \mathbf{C}_{b(\sigma)}^{b(t_k)}\,\mathbf{f}^b d\sigma d\tau\right]$$

$$(35)$$

where the first integral has been given as in (29)-(30) and the second integral is approximated using the two-sample correction by (see Appendix B for details)

$$\int_{t_k}^{t_{k+1}}\int_{t_k}^\tau \mathbf{C}_{b(\sigma)}^{b(t_k)}\,\mathbf{f}^b d\sigma d\tau=\frac{T}{30}\left(25\Delta\mathbf{v}_1+5\Delta\mathbf{v}_2+12\Delta\boldsymbol{\theta}_1\times\Delta\mathbf{v}_1+8\Delta\boldsymbol{\theta}_1\times\Delta\mathbf{v}_2+2\Delta\mathbf{v}_1\times\Delta\boldsymbol{\theta}_2+2\Delta\boldsymbol{\theta}_2\times\Delta\mathbf{v}_2\right) \quad (36)$$

### C. In-flight Alignment Algorithms Derived from Velocity/ Position Integration Formulae

Two in-flight alignment algorithms are now constructed respectively based on the velocity integration formula (11) or the position integration formula (20). Ignoring the subscripts that indicates velocity or position, the alignment is to solve the initial attitude matrix from

$$\mathbf{C}_b^n\left(0\right)\boldsymbol{\alpha}\left(t_M\right)=\boldsymbol{\beta}\left(t_M\right)\quad\left(M=1,2,\ldots\right) \tag{37}$$

where $\boldsymbol{\beta}$, given as (28) or (34), is a known time-varying vector as a function of the aided velocity and position, and $\boldsymbol{\alpha}$, given as (29) or (35), is a known time-varying vector as a function of gyroscope and accelerometer measurements. As shown in [10], (37) can be solved by the optimization-based method using the unit attitude quaternion parameter. It should be highlighted that (11) or (20) is an integral equation at any time, in contrast to the differential equation in (6) in [10]. This spares the calculation of the derivative of the velocity that is necessitated if (6) in [10] was otherwise used, and the integrals involved can also significantly depress the noise in the aided velocity/position measurements.

Specifically, the four-element unit quaternion $\mathbf{q}=\begin{bmatrix}s & \boldsymbol{\eta}^T\end{bmatrix}^T$, where $s$ is the scalar part and $\boldsymbol{\eta}$ is the vector part, is used to encode the initial body attitude matrix $\mathbf{C}_n^b\left(0\right)$. The relationship between these two rotation parameters is (Note a sign typo in (9) of [10])

$$\mathbf{C}_n^b\left(0\right)=\left(s^2-\boldsymbol{\eta}^T\boldsymbol{\eta}\right)\mathbf{I}+2\boldsymbol{\eta}\boldsymbol{\eta}^T-2s\left(\boldsymbol{\eta}\times\right) \tag{38}$$

Define the quaternion multiplication matrices by

$$\left[\mathbf{q}\right]^{+} \triangleq \begin{bmatrix} s & -\boldsymbol{\eta}^{T} \\ \boldsymbol{\eta} & s\,I + (\boldsymbol{\eta}\times) \end{bmatrix}, \quad \left[\mathbf{q}\right]^{-} \triangleq \begin{bmatrix} s & -\boldsymbol{\eta}^{T} \\ \boldsymbol{\eta} & s\,I - (\boldsymbol{\eta}\times) \end{bmatrix} \tag{39}$$

Then (37) is equivalent to $\left( \left[ \boldsymbol{\beta}\left(\overset{+}{t}_{M}\right) \right] - \left[ \boldsymbol{\alpha}\left(\overset{-}{t}_{M}\right) \right] \right)\mathbf{q} = 0$, and the determination of the attitude quaternion can be posed as a constrained optimization [10]

$$\min_{\mathbf{q}} \mathbf{q}^{T}\mathbf{K}\mathbf{q}, \text{ subject to } \mathbf{q}^{T}\mathbf{q} = 1 \tag{40}$$

where the $4 \times 4$ real symmetric matrix

$$\mathbf{K} \triangleq \sum_{M} \left( \left[ \boldsymbol{\beta}\left(\overset{+}{t}_{M}\right) \right] - \left[ \boldsymbol{\alpha}\left(\overset{-}{t}_{M}\right) \right] \right)^{T} \left( \left[ \boldsymbol{\beta}\left(\overset{+}{t}_{M}\right) \right] - \left[ \boldsymbol{\alpha}\left(\overset{-}{t}_{M}\right) \right] \right) \tag{41}$$

It can be proved that the optimal quaternion is exactly the normalized eigenvector of $\mathbf{K}$ belonging to the smallest eigenvalue [10, 11].

For more clarity, the in-flight algorithm derived from the velocity integration formula (IFA-VIF) is listed in Table I and the in-flight algorithm derived from the position integration formula (IFA-PIF) is listed in Table II.

## V. Simulation Results

This section is devoted to numerically examine the in-flight algorithms proposed in this paper. We carried out scenarios with oscillating attitude and translation to simulate large motion maneuvers. The SINS under investigation is located at medium latitude $30°$ that is equipped with a triad of gyroscopes (drift $0.01°/h$, noise $0.1°/h/\sqrt{\text{Hz}}$) and accelerometers (bias $50\mu g$, noise $500\mu g/\sqrt{\text{Hz}}$). The SINS sampling rate is 100 $Hz$. Figures 1-2 show the true attitude and velocity profiles for the first 300 seconds. The aided velocity and position are provided by the external absolute sensor, such as GPS, displaced from the SINS by a lever arm $\mathbf{l}^{b} = \begin{bmatrix} 1 & 1 & 1 \end{bmatrix}^{T} m$ in the SINS body frame. Existence of the GPS lever arm is common for applications, whose effect cannot be neglected in large angular motions. White velocity noise (standard variance 0.1 $m/s$) and position noise (standard variance 2 $m$) are added to reflect the single-point GPS measurement quality. The incremental update interval $T = 0.02\,s$.

We first carry out the ideal case with perfect sensors, i.e., no SINS/GPS measurement errors and no GPS lever arm. As shown in Fig. 3, both IFA-VIF and IFA-PIF produce the correct attitude angles with negligible errors, which validates the correctness of the above analysis and the implemented algorithms. Next we evaluate the two algorithms by implementing 100 Monte Carlo runs. Figure 4 plots the mean alignment angle errors for IFA-VIF and IFA-PIF, as well as their respective 3 $\sigma$ envelopes. It appears that both IFA-VIF and IFA-PIF have error lumps due to the lever arm presence as shown below. On the other hand, the IFA-VIF estimate has shorter stabilizing time but about one time larger scattering scope. For example, the IFA-VIF angle errors (3 $\sigma$) reduce to $\begin{bmatrix} -0.006 \pm 0.004 & 0.021 \pm 0.227 & 0.003 \pm 0.005 \end{bmatrix}^T$ in degree at 300s, in contrast to the IFA-PIF angle errors $\begin{bmatrix} -0.006 \pm 0.003 & -0.012 \pm 0.129 & 0.005 \pm 0.003 \end{bmatrix}^T$ in degree. In order to examine the lever arm effect, the GPS-error-contaminated SINS velocity and position are used as inputs to the algorithms, which corresponds to the non-practical case with zero GPS lever arm. The mean alignment angle errors of IFA-VIF, with and without the lever arm, are presented in Fig. 5. The presence of the lever arm leads to obvious lump peaks at about 30s and biased angle estimates. Similar phenomenon can be observed for IFA-PIF as shown in (Fig. 6.

## VI. FIELD TEST RESULTS

A flight test data was collected on board the Cessna 208 aircraft in 2010, as shown in Fig. 7. A navigation-grade SINS is fixed inside the cabin, with the GPS antenna installed outside the cabin on the top of the aircraft. The SINS-GPS lever arm is about 2.33 m (forward), 0.35 m (right), and 1.35 m (upward), roughly measured by a ruler. The SINS/GPS integrated navigation result is used as the reference data to compare with, in which the GPS antenna's lever arm has been estimated and eliminated to great extent [19]. The SINS/GPS flight trajectory is shown in Fig. 8, overlapped by three chosen 100-seconds test segments with large maneuvers: the ascending segment (S1), the turning segment (S2) and the descending segment (S3).

The GPS raw measurements (2 $Hz$) were linearly interpolated to obtain the velocity and position at both ends of the update interval $T = 0.02s$. The velocity profile for the test segment S1 is plotted in Fig. 9. The IFA-VIF and IFA-PIF estimate results against the reference true values are given in Fig. 10. The IFA-VIF heading error reduces to about 5 degree in 10 seconds and the two IFA-VIF level angles reduces to within 1 degree in 5 seconds. In comparison, IFA-PIF heading error reduces to about 13 degree in 10 seconds and about 5 degree in 20 seconds (the sudden jump at 60s is due to the Euler angle's ambiguity) and the two IFA-PIF level angles reduces to within 2 degree in 5 seconds. Figures 11-12 respectively provide the velocity profile and attitude estimate for the test segment

S2, and Figs. 13-14 respectively provide the velocity profile and attitude estimate for the test segment S3. As per S2-S3 segments, the yaw angle errors are below 3 degree for IFA-VIF and below 5 degree for IFA-PIF in 20 seconds, while the level angle errors are below 2 degree in 5 seconds for both IFA-VIF and IFA-PIF. The angle errors as a function of elapsed time are summarized in Table III for three test segments.

To order to examine the lever arm effect, the in-flight alignment using the INS/GPS reference velocity and position was carried out. Figure 15-16 plot the obtained yaw errors, by IFA-VIF and IFA-PIF, respectively, for three test segments against those using GPS raw measurements. The attitude error summary is also listed in Table III for comparison. With the lever arm eliminated, the IFA-VIF heading error is less than 1 degree in 10s, less than 0.3 degree in 20s, and the IFA-VIF level angle errors are less than 0.1 degree in 10 seconds, while the IFA-PIF heading error is less than 5 degree in 10s, less than 2 degree in 20s, and the IFA-PIF level angle errors are less than 0.25 degree in 10 seconds. It shows that the IFA-VIF performs better than IFA-PIF in this flight test.

The remarkable improvement (over ten times for IFA-VIF) tells us that the lever arm effect should be seriously considered or avoided for the in-flight alignment. It can be minimized by placing the GPS antenna as close to the SINS as possible, or avoiding quick turns during alignment. Note that the lever arm, even known precisely in practice, could not be easily compensated because it inversely needs the knowledge of the attitude that we are trying to solve.

## VII. Conclusions

A good attitude initialization is very important for rapid and accurate SINS fine alignment. It is generally a difficult problem to acquire a rough initial angle when the carrier is moving or maneuvering. This paper proposes an optimization-based in-flight coarse alignment approach based on the velocity/position integration formulae that are derived from the traditional navigation rate equation by integration manipulations. The integrals involved are delicately handled so as to reduce the calculation errors caused by maneuvers as much as possible and two recursive alignment algorithms are designed. Simulations and flight test data are used to evaluate the two alignment algorithms. The results show that, with the GPS lever arm well handled, the SINS heading can be potentially aligned up to one degree accuracy in ten seconds. The algorithms proposed in this paper could be useful to many applications that require aligning the INS on the run. In the future work, we will test the algorithms further on low quality SINS and try to handle the sensor bias using the proposed optimization approach.

APPENDIX

*A. Two-sample Approximation of Integral in (29)*

Assume the body angular velocity and the specific force are both approximate linear forms as

$$\boldsymbol{\omega}_{ib}^b = t\,\mathbf{a}_\omega + \mathbf{b}_\omega$$
$$\mathbf{f}^b = t\,\mathbf{a}_f + \mathbf{b}_f \tag{42}$$

where $\mathbf{a}_\omega$, $\mathbf{b}_\omega$, $\mathbf{a}_f$ and $\mathbf{b}_f$ are the appropriate coefficient vectors. Then we have for the incremental angle

$$\Delta\boldsymbol{\theta}_1 = \int_0^{T/2} \boldsymbol{\omega}_{ib}^b dt = \int_0^{T/2} t\,\mathbf{a}_\omega + \mathbf{b}_\omega dt = \frac{T^2}{8}\mathbf{a}_\omega + \frac{T}{2}\mathbf{b}_\omega$$
$$\Delta\boldsymbol{\theta}_1 + \Delta\boldsymbol{\theta}_2 = \int_0^{T} \boldsymbol{\omega}_{ib}^b dt = \int_0^{T} t\,\mathbf{a}_\omega + \mathbf{b}_\omega dt = \frac{T^2}{2}\mathbf{a}_\omega + T\mathbf{b}_\omega \tag{43}$$

from which the coefficient vectors are solved as

$$T^2\mathbf{a}_\omega = 4\left(\Delta\boldsymbol{\theta}_2 - \Delta\boldsymbol{\theta}_1\right)$$
$$T\mathbf{b}_\omega = 3\Delta\boldsymbol{\theta}_1 - \Delta\boldsymbol{\theta}_2 \tag{44}$$

Similarly for the incremental velocity, we get

$$T^2\mathbf{a}_f = 4\left(\Delta\mathbf{v}_2 - \Delta\mathbf{v}_1\right)$$
$$T\mathbf{b}_f = 3\Delta\mathbf{v}_1 - \Delta\mathbf{v}_2 \tag{45}$$

Then the integral in (29) is approximated by

$$\begin{aligned}
\int_{t_k}^{t_{k+1}} &\left(\mathbf{I} + \left(\int_{t_k}^{t} \boldsymbol{\omega}_{ib}^b d\tau\right)\times\right)\mathbf{f}^b dt = \int_{t_k}^{t_{k+1}} \left[\mathbf{I} + \left(\int_{t_k}^{t} (\tau - t_k)\mathbf{a}_\omega + \mathbf{b}_\omega d\tau\right)\times\right]\left[(t - t_k)\mathbf{a}_f + \mathbf{b}_f\right]dt \\
&= \int_{t_k}^{t_{k+1}} (t - t_k)\,\mathbf{a}_f + \mathbf{b}_f + \left(\frac{1}{2}(t - t_k)^2\,\mathbf{a}_\omega + (t - t_k)\mathbf{b}_\omega\right)\times\left(t\,\mathbf{a}_f + \mathbf{b}_f\right)dt \\
&= \int_{t_k}^{t_{k+1}} \mathbf{b}_f + (t - t_k)\left(\mathbf{a}_f + \mathbf{b}_\omega\times\mathbf{b}_f\right) + (t - t_k)^2\left(\frac{1}{2}\mathbf{a}_\omega\times\mathbf{b}_f + \mathbf{b}_\omega\times\mathbf{a}_f\right) + \frac{1}{2}(t - t_k)^3\,\mathbf{a}_\omega\times\mathbf{a}_f\,dt \\
&= T\,\mathbf{b}_f + \frac{1}{2}T^2\left(\mathbf{a}_f + \mathbf{b}_\omega\times\mathbf{b}_f\right) + \frac{1}{3}T^3\left(\frac{1}{2}\mathbf{a}_\omega\times\mathbf{b}_f + \mathbf{b}_\omega\times\mathbf{a}_f\right) + \frac{1}{8}T^4\mathbf{a}_\omega\times\mathbf{a}_f \\
&= \Delta\mathbf{v}_1 + \Delta\mathbf{v}_2 + \frac{1}{2}\left(\Delta\boldsymbol{\theta}_1 + \Delta\boldsymbol{\theta}_2\right)\times\left(\Delta\mathbf{v}_1 + \Delta\mathbf{v}_2\right) + \frac{2}{3}\left(\Delta\boldsymbol{\theta}_1\times\Delta\mathbf{v}_2 + \Delta\mathbf{v}_1\times\Delta\boldsymbol{\theta}_2\right)
\end{aligned} \tag{46}$$

*B. Two-sample Approximation of Integral in (35)*

With (46), we have

$$\int_{t_k}^{t_{k+1}} \int_{t_k}^{\tau} \mathbf{C}_{b(\sigma)}^{b(t_k)} \mathbf{f}^b d\sigma d\tau$$

$$= \int_{t_k}^{t_{k+1}} (\tau - t_k) \mathbf{b}_f + \frac{1}{2}(\tau - t_k)^2 (\mathbf{a}_f + \mathbf{b}_\omega \times \mathbf{b}_f) + \frac{1}{3}(\tau - t_k)^3 \left(\frac{1}{2}\mathbf{a}_\omega \times \mathbf{b}_f + \mathbf{b}_\omega \times \mathbf{a}_f\right) + \frac{1}{8}(\tau - t_k)^4 \mathbf{a}_\omega \times \mathbf{a}_f d\tau$$

$$= \frac{1}{2}T^2 \mathbf{b}_f + \frac{1}{6}T^3 (\mathbf{a}_f + \mathbf{b}_\omega \times \mathbf{b}_f) + \frac{1}{12}T^4 \left(\frac{1}{2}\mathbf{a}_\omega \times \mathbf{b}_f + \mathbf{b}_\omega \times \mathbf{a}_f\right) + \frac{1}{40}T^5 \mathbf{a}_\omega \times \mathbf{a}_f$$

$$= \frac{T}{30}\left(25\Delta\mathbf{v}_1 + 5\Delta\mathbf{v}_2 + 12\Delta\boldsymbol{\theta}_1 \times \Delta\mathbf{v}_1 + 8\Delta\boldsymbol{\theta}_1 \times \Delta\mathbf{v}_2 + 2\Delta\mathbf{v}_1 \times \Delta\boldsymbol{\theta}_2 + 2\Delta\boldsymbol{\theta}_2 \times \Delta\mathbf{v}_2\right)$$

$$(47)$$

*C. Derivation of $\boldsymbol{\varphi}_b$*

The body rotation vector can be approximated by

$$\boldsymbol{\varphi}_b \approx \int_{t_k}^{t_{k+1}} \left(\boldsymbol{\omega}_{ib}^b + \frac{1}{2}\left(\int_{t_k}^{t} \boldsymbol{\omega}_{ib}^b d\tau\right) \times \boldsymbol{\omega}_{ib}^b\right) dt$$

$$= \Delta\boldsymbol{\theta}_1 + \Delta\boldsymbol{\theta}_2 + \frac{1}{2}\int_{t_k}^{t_{k+1}} \left(\frac{1}{2}(t - t_k)^2 \mathbf{a}_\omega + (t - t_k)\mathbf{b}_\omega\right) \times \left((t - t_k)\mathbf{a}_\omega + \mathbf{b}_\omega\right) dt$$

$$= \Delta\boldsymbol{\theta}_1 + \Delta\boldsymbol{\theta}_2 - \frac{T^3}{12}\mathbf{a}_\omega \times \mathbf{b}_\omega$$

$$= \Delta\boldsymbol{\theta}_1 + \Delta\boldsymbol{\theta}_2 + \frac{2}{3}\Delta\boldsymbol{\theta}_1 \times \Delta\boldsymbol{\theta}_2$$

$$(48)$$

## ACKNOWLEDGEMENTS

Thanks to Dr. Kaidong Zhang, Yangming Huang and Shaokun Cai for providing the flight test data.

Table I. In-flight Algorithm Derived from Velocity Integration Formula (IFA-VIF)

| | |
|---|---|
| ***Initialization:*** | *Set* $M = 0$, $\boldsymbol{\alpha}_v(0) = \boldsymbol{\beta}'_v(0) = \mathbf{0}_{3\times 1}$ *and* $\mathbf{K}(0) = \mathbf{0}_{4\times 4}$. |
| ***Step 1:*** | *Set* $M = M + 1$; |
| ***Step 2:*** *(see (5))* | *Update* $\mathbf{C}_{n(t_{M-1})}^{n(0)}$ *to* $\mathbf{C}_{n(t_M)}^{n(0)}$ *using* $\boldsymbol{\omega}_{in}^n$ *by*<br><br>$\boldsymbol{\varphi}_n = T\boldsymbol{\omega}_{in}^n$, $C_{n(t_M)}^{n(t_{M-1})} = I + \dfrac{\sin\left(\left\|\boldsymbol{\varphi}_n\right\|\right)}{\left\|\boldsymbol{\varphi}_n\right\|}\boldsymbol{\varphi}_n \times + \dfrac{1-\cos\left(\left\|\boldsymbol{\varphi}_n\right\|\right)}{\left\|\boldsymbol{\varphi}_n\right\|^2}(\boldsymbol{\varphi}_n \times)^2$, $\mathbf{C}_{n(t_M)}^{n(0)} = \mathbf{C}_{n(t_{M-1})}^{n(0)} C_{n(t_M)}^{n(t_{M-1})}$<br><br>*Update* $\mathbf{C}_{b(t_{M-1})}^{b(0)}$ *to* $\mathbf{C}_{b(t_M)}^{b(0)}$ *using the gyroscope-measured angular increments by*[*]<br><br>$\boldsymbol{\varphi}_b = \Delta\boldsymbol{\theta}_1 + \Delta\boldsymbol{\theta}_2 + \dfrac{2}{3}\Delta\boldsymbol{\theta}_1 \times \Delta\boldsymbol{\theta}_2$, $C_{b(t_M)}^{b(t_{M-1})} = I + \dfrac{\sin\left(\left\|\boldsymbol{\varphi}_b\right\|\right)}{\left\|\boldsymbol{\varphi}_b\right\|}\boldsymbol{\varphi}_b \times + \dfrac{1-\cos\left(\left\|\boldsymbol{\varphi}_b\right\|\right)}{\left\|\boldsymbol{\varphi}_b\right\|^2}(\boldsymbol{\varphi}_b \times)^2$, $\mathbf{C}_{b(t_M)}^{b(0)} = \mathbf{C}_{b(t_{M-1})}^{b(0)} C_{b(t_M)}^{b(t_{M-1})}$: |
| ***Step 3:*** *(see (29)-(30))* | *Compute* $\boldsymbol{\alpha}_v(t_M)$ *using the gyroscope/accelerometer outputs by*<br><br>$\boldsymbol{\alpha}_v(t_M) = \boldsymbol{\alpha}_v(t_{M-1}) + \mathbf{C}_{b(t_{M-1})}^{b(0)}\left[\Delta\mathbf{v}_1 + \Delta\mathbf{v}_2 + \dfrac{1}{2}(\Delta\boldsymbol{\theta}_1 + \Delta\boldsymbol{\theta}_2) \times (\Delta\mathbf{v}_1 + \Delta\mathbf{v}_2) + \dfrac{2}{3}(\Delta\boldsymbol{\theta}_1 \times \Delta\mathbf{v}_2 + \Delta\mathbf{v}_1 \times \Delta\boldsymbol{\theta}_2)\right]$; |
| ***Step 4:*** *(see (28))* | *Compute* $\boldsymbol{\beta}_v(t_M)$ *using the aided velocity and position by*<br><br>$\boldsymbol{\beta}'_v(t_M) = \boldsymbol{\beta}'_v(t_{M-1}) + \mathbf{C}_{n(t_{M-1})}^{n(0)}\left[\left(\dfrac{T}{2}\mathbf{I} + \dfrac{T^2}{6}\boldsymbol{\omega}_{in}^n\times\right)\boldsymbol{\omega}_{ie}^n \times \mathbf{v}^n(t_{M-1}) + \left(\dfrac{T}{2}\mathbf{I} + \dfrac{T^2}{3}\boldsymbol{\omega}_{in}^n\times\right)\boldsymbol{\omega}_{ie}^n \times \mathbf{v}^n(t_M) - \left(T\mathbf{I} + \dfrac{T^2}{2}\boldsymbol{\omega}_{in}^n\times\right)\mathbf{g}^n\right]$<br><br>$\boldsymbol{\beta}_v(t_M) = \mathbf{C}_{n(t_M)}^{n(0)}\mathbf{v}^n(0) - \mathbf{v}^n(t_M) + \boldsymbol{\beta}'_v(t_M)$: |
| ***Step 5:*** *(see (39) and (41))* | *Compute* $\mathbf{K}(t_M) = \mathbf{K}(t_{M-1}) + \left(\left[\boldsymbol{\beta}_v(t_M)\right] - \left[\boldsymbol{\alpha}_v(t_M)\right]\right)^T\left(\left[\boldsymbol{\beta}_v(t_M)\right] - \left[\boldsymbol{\alpha}_v(t_M)\right]\right)$; |
| ***Step 6:*** | *Determine the initial attitude quaternion* $\mathbf{q}$ *by calculating the normalized eigenvector of* $\mathbf{K}(t_M)$ *belonging to the smallest eigenvalue;* |
| ***Step 7:*** *(see (38) and (4))* | *Obtain the attitude matrix at current time;* |
| ***Step 8:*** | *Go to Step 1 until the end.* |

[*]: See Appendix C for the derivation of $\boldsymbol{\varphi}_b$.

Table II. In-flight Algorithm Derived from Position Integration Formula (IFA-PIF)

| | |
|---|---|
| ***Initialization:*** | *Set* $M = 0$, $\boldsymbol{\alpha}_p(0) = \mathbf{r}^n(0) = \mathbf{0}_{3\times 1}$, $\mathbf{u}_r(0) = \mathbf{u}_v(0) = \mathbf{u}_g(0) = \mathbf{0}_{3\times 1}$ *and* $\mathbf{K}(0) = \mathbf{0}_{4\times 4}$. |
| ***Step 1:*** | *Set* $M = M + 1$; |
| ***Step 2:*** | *See Step 2 in Table I;* |
| ***Step 3:*** *(see* (35)-(36)*)* | *Compute* $\boldsymbol{\alpha}_p(t_M)$ *using the gyroscope/accelerometer outputs by*  $$\boldsymbol{\alpha}_p(t_M) = \boldsymbol{\alpha}_p(t_{M-1}) + T\sum_{m=0}^{M-2}\mathbf{C}_{b(t_m)}^{n(0)}\left(\Delta\mathbf{v}_1 + \Delta\mathbf{v}_2 + \frac{1}{2}(\Delta\boldsymbol{\theta}_1 + \Delta\boldsymbol{\theta}_2)\times(\Delta\mathbf{v}_1 + \Delta\mathbf{v}_2) + \frac{2}{3}(\Delta\boldsymbol{\theta}_1\times\Delta\mathbf{v}_2 + \Delta\mathbf{v}_1\times\Delta\boldsymbol{\theta}_2)\right)$$ $$+\mathbf{C}_{b(t_{M-1})}^{n(0)}\left[\frac{T}{30}\left(25\Delta\mathbf{v}_1 + 5\Delta\mathbf{v}_2 + 12\Delta\boldsymbol{\theta}_1\times\Delta\mathbf{v}_1 + 8\Delta\boldsymbol{\theta}_1\times\Delta\mathbf{v}_2 + 2\Delta\mathbf{v}_1\times\Delta\boldsymbol{\theta}_2 + 2\Delta\boldsymbol{\theta}_2\times\Delta\mathbf{v}_2\right)\right]$$ ; |
| ***Step 4:*** *(see* (34)*)* | *Compute* $\boldsymbol{\beta}_p(t_M)$ *using the aided velocity and position by*  $$\mathbf{u}_r(t_M) = \mathbf{u}_r(t_{M-1}) + \mathbf{C}_{n(t_{M-1})}^{n(0)}\left[\left(\frac{T}{2}I + \frac{T^2}{6}\boldsymbol{\omega}_{in}^n\times\right)\mathbf{v}^n(t_k) + \left(\frac{T}{2}I + \frac{T^2}{3}\boldsymbol{\omega}_{in}^n\times\right)\mathbf{v}^n(t_{k+1})\right]$$ $$\mathbf{u}_v(t_M) = \mathbf{u}_v(t_{M-1}) + \mathbf{C}_{n(t_{M-1})}^{n(0)}\left[\left(\frac{T^2}{3}\mathbf{I} + \frac{T^3}{12}\boldsymbol{\omega}_{in}^n\times\right)\boldsymbol{\omega}_{ie}^n\times\mathbf{v}^n(t_{M-1}) + \left(\frac{T^2}{6}\mathbf{I} + \frac{T^3}{12}\boldsymbol{\omega}_{in}^n\times\right)\boldsymbol{\omega}_{ie}^n\times\mathbf{v}^n(t_M)\right]$$ $$+T\sum_{m=0}^{M-2}\mathbf{C}_{n(t_m)}^{n(0)}\left[\left(\frac{T}{2}\mathbf{I} + \frac{T^2}{6}\boldsymbol{\omega}_{in}^n\times\right)\boldsymbol{\omega}_{ie}^n\times\mathbf{v}^n(t_m) + \left(\frac{T}{2}\mathbf{I} + \frac{T^2}{2}\boldsymbol{\omega}_{in}^n\times\right)\boldsymbol{\omega}_{ie}^n\times\mathbf{v}^n(t_{m+1})\right]$$ $$\mathbf{u}_g(t_M) = \mathbf{u}_g(t_{M-1}) + \mathbf{C}_{n(t_{M-1})}^{n(0)}\left(\frac{T^2}{6}\mathbf{I} + \frac{T^3}{6}\boldsymbol{\omega}_{in}^n\times\right)\mathbf{g}^n + T\sum_{m=0}^{M-2}\mathbf{C}_{n(t_m)}^{n(0)}\left(T\mathbf{I} + \frac{T^2}{2}\boldsymbol{\omega}_{in}^n\times\right)\mathbf{g}^n$$ $$\boldsymbol{\beta}_p(t_M) = \mathbf{u}_r(t_M) - t_M\mathbf{v}^n(0) + \mathbf{u}_v(t_M) - \mathbf{u}_g(t_M) ;$$ |
| ***Step 5:*** *(see* (39) *and* (41)*)* | *Compute* $\mathbf{K}(t_M) = \mathbf{K}(t_{M-1}) + \left(\left[\boldsymbol{\beta}_p(t_M)\right] - \left[\boldsymbol{\alpha}_p(t_M)\right]\right)^T\left(\left[\boldsymbol{\beta}_p(t_M)\right] - \left[\boldsymbol{\alpha}_p(t_M)\right]\right);$ |
| ***Step 6-7:*** | *See Step 6-7 in Table I;* |
| ***Step 8:*** | *Go to Step 1 until the end.* |

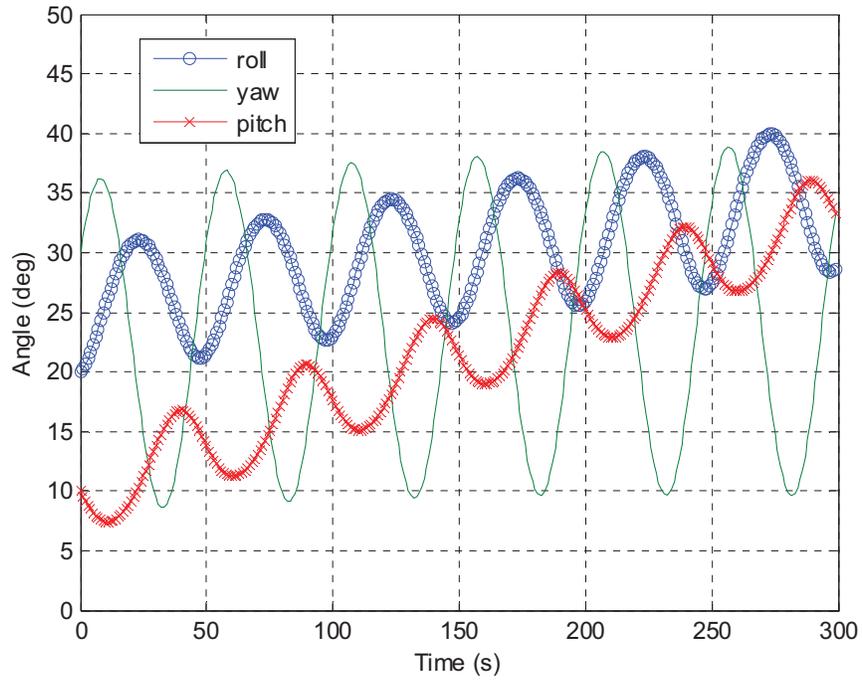

Figure 1. True attitude angle profile for simulated scenarios.

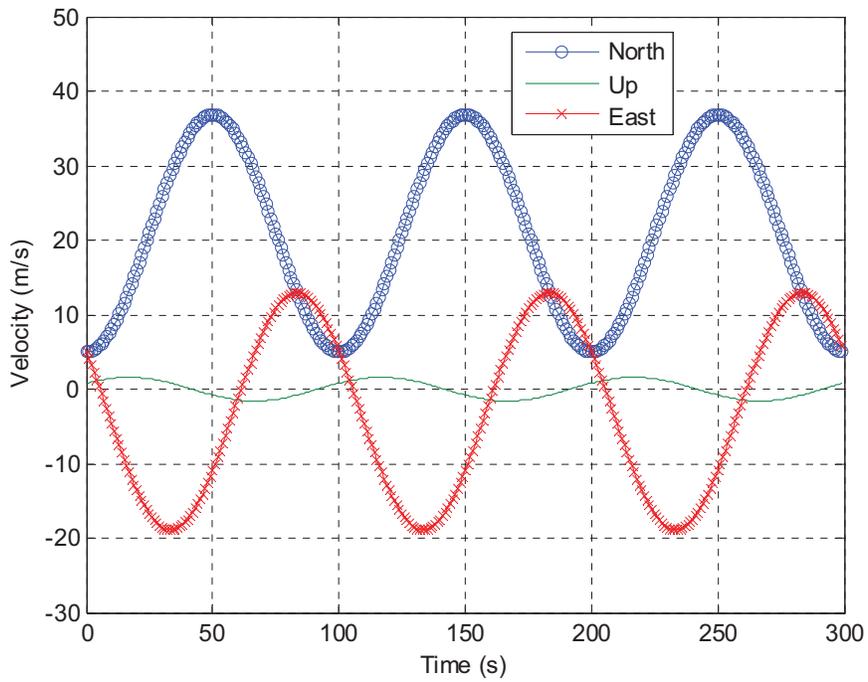

Figure 2. True velocity profile for simulated scenarios.

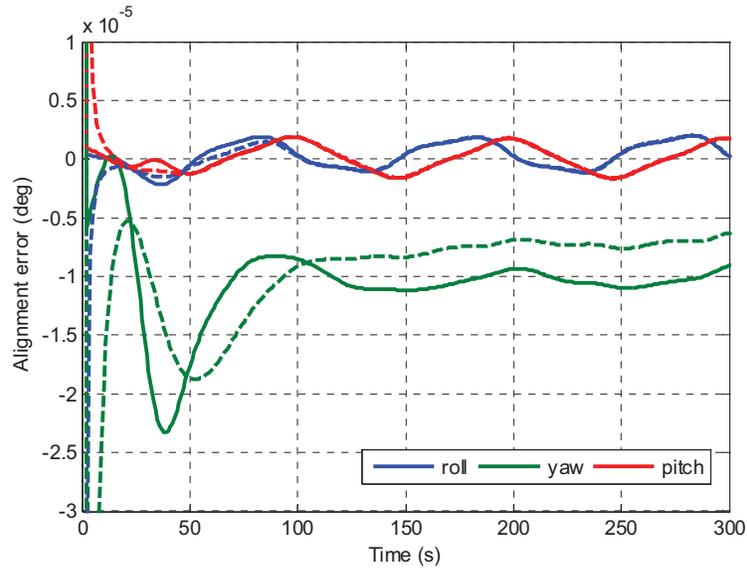

Figure 3. Alignment errors for ideal case with perfect sensors (IFA-VIF: solid lines; IFA-PIF: dashed lines).

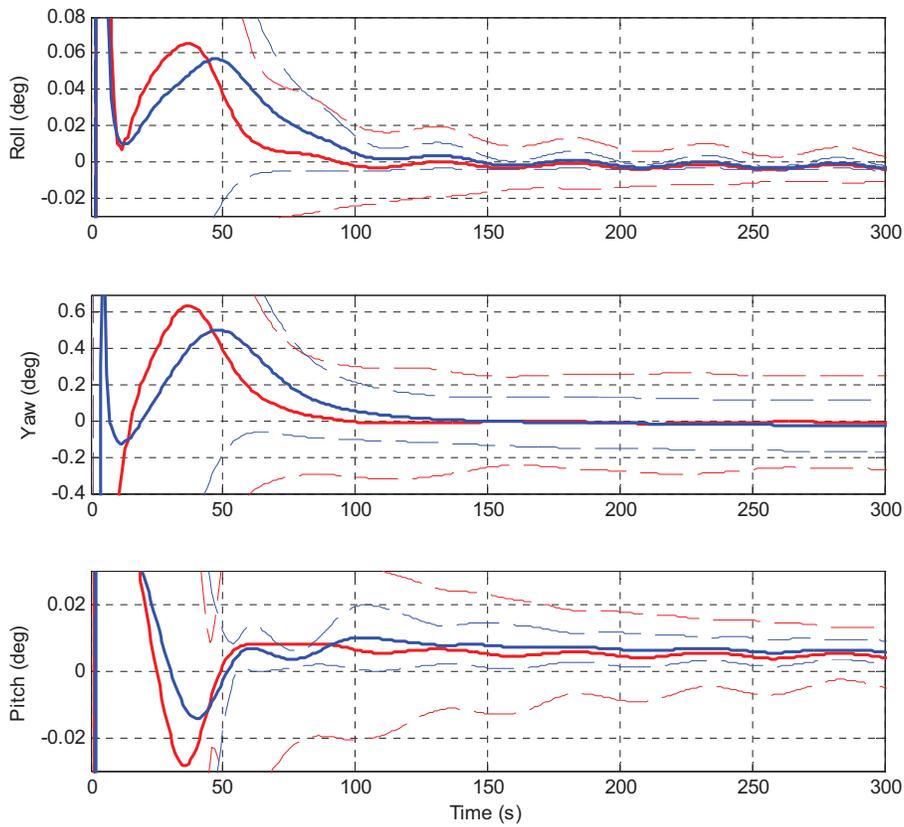

Figure 4. Mean alignment angle errors and their $3\sigma$ envelopes (IFA-VIF: red lines; IFA-PIF: blue lines).

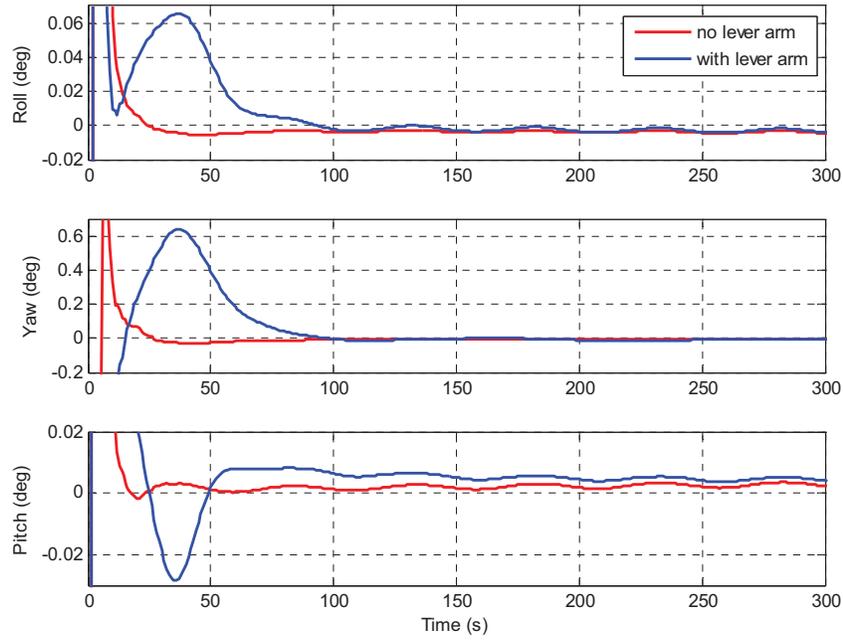

Figure 5. Mean alignment angle errors for IFA-VIF with and without the lever arm

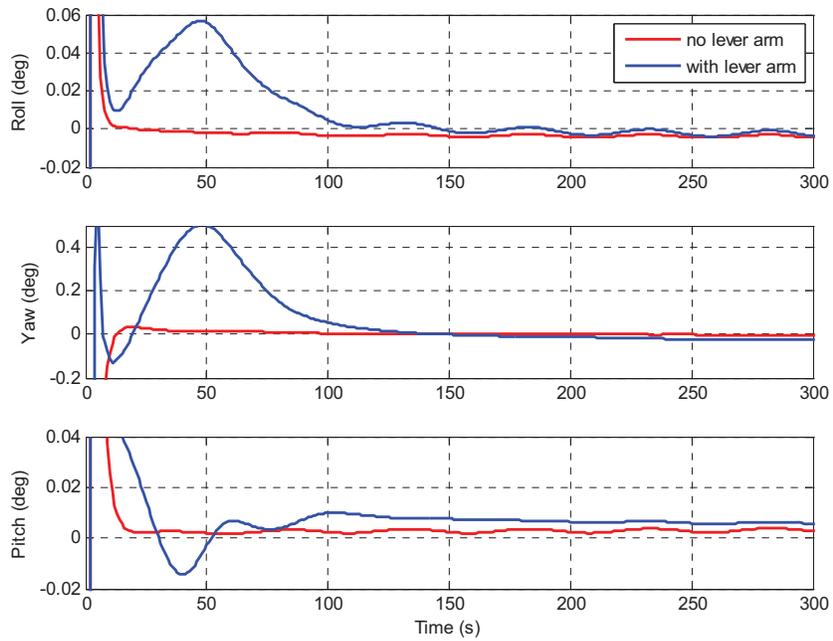

Figure 6. Mean alignment angle errors for IFA-PIF with and without the lever arm

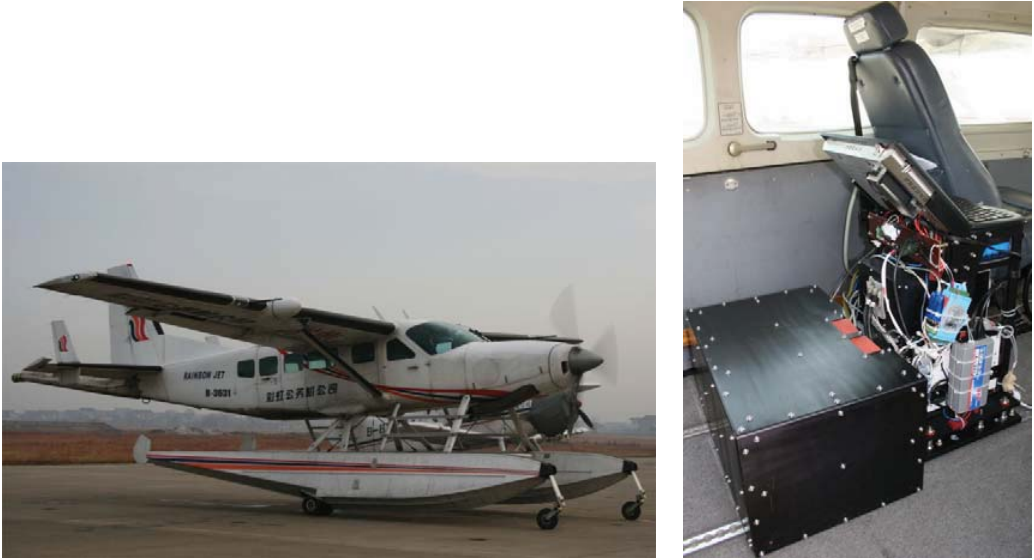

Figure 7. Cessna 208 test aircraft (left) and a snapshot of the SINS (right).

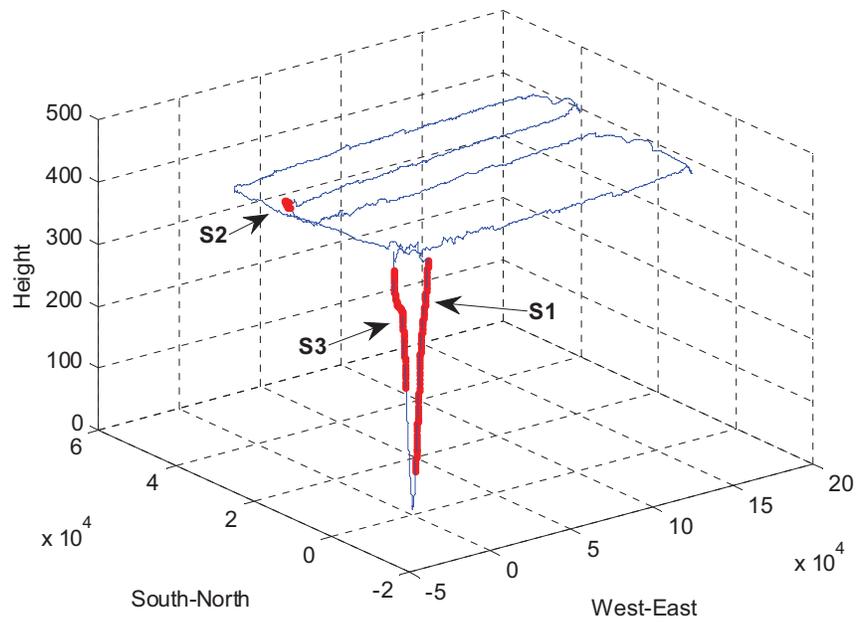

Figure 8. Flight trajectory (blue line) and three test segments (red dot). Unit: meter.

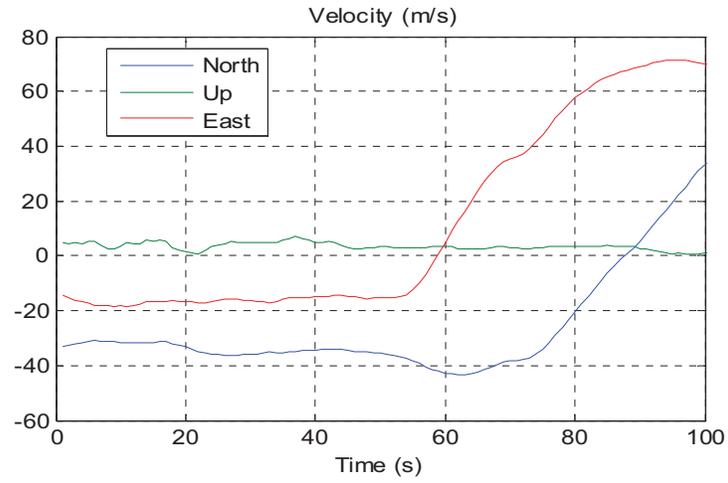

Figure 9. Velocity profile for S1.

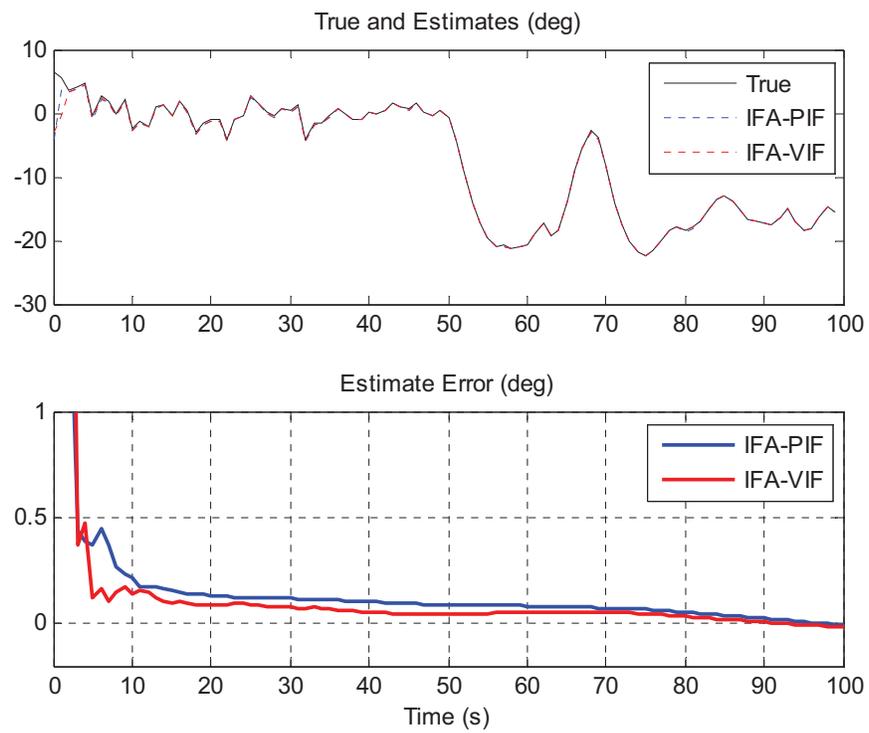

Figure 10a. Roll angle estimate and error for S1.

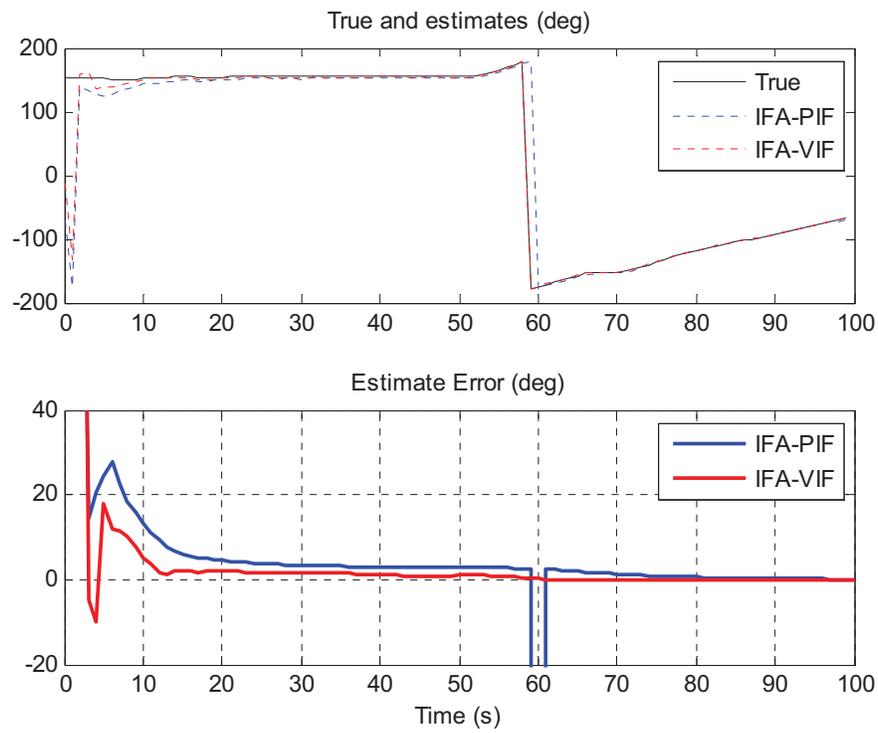

Figure 10b. Yaw angle estimate and error for S1.

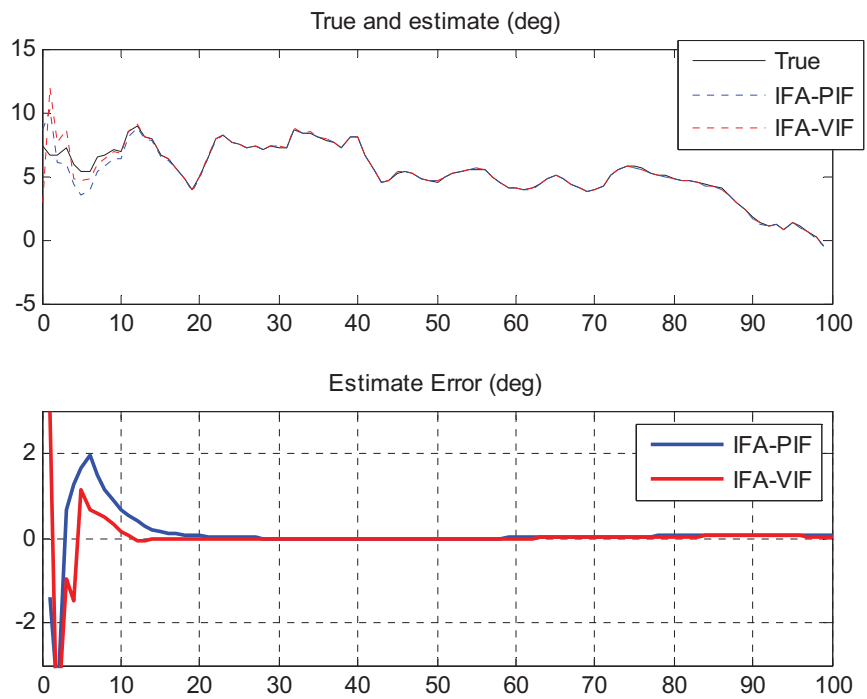

Figure 10b. Pitch angle estimate and error for S1.

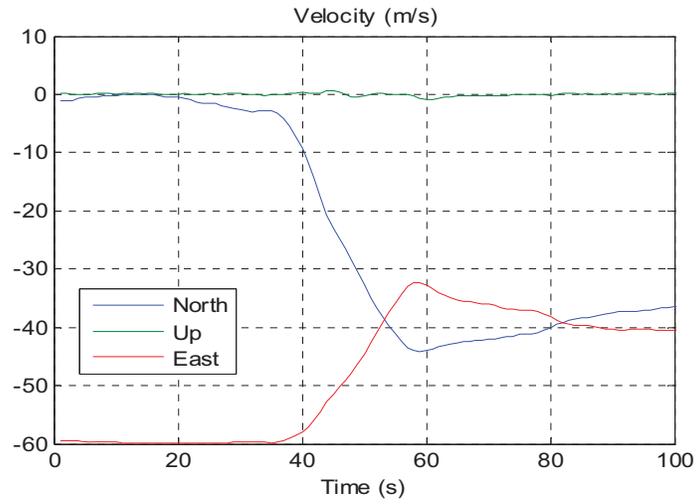

Figure 11. Velocity profile for S2.

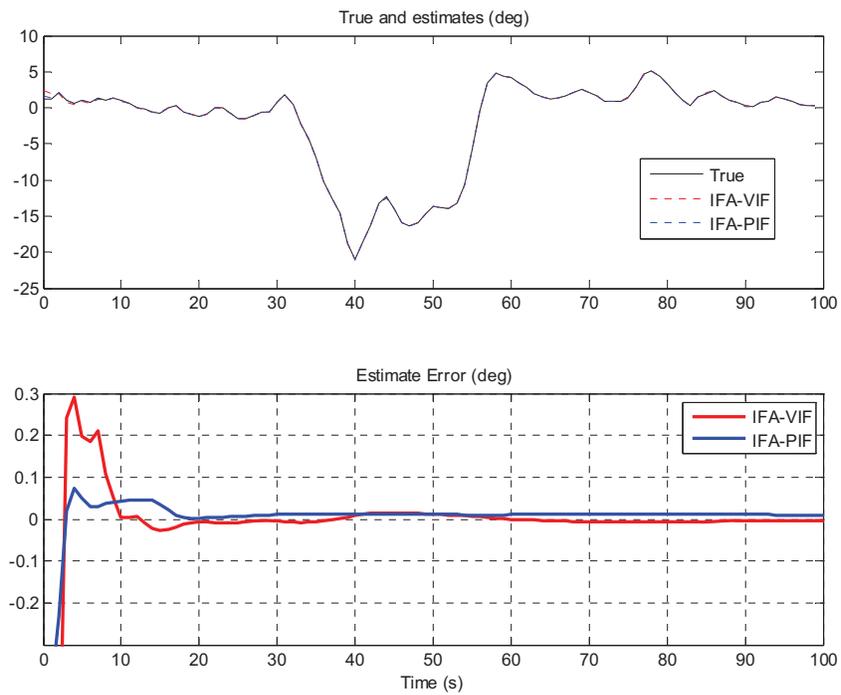

Figure 12a. Roll angle estimate and error for S2.

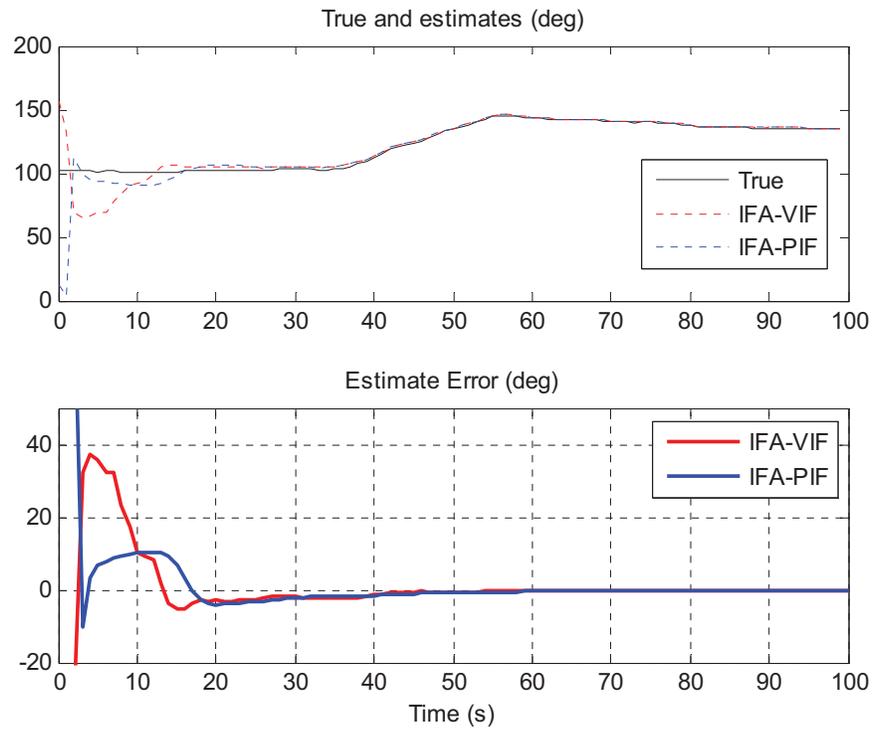

Figure 12b. Yaw angle estimate and error for S2.

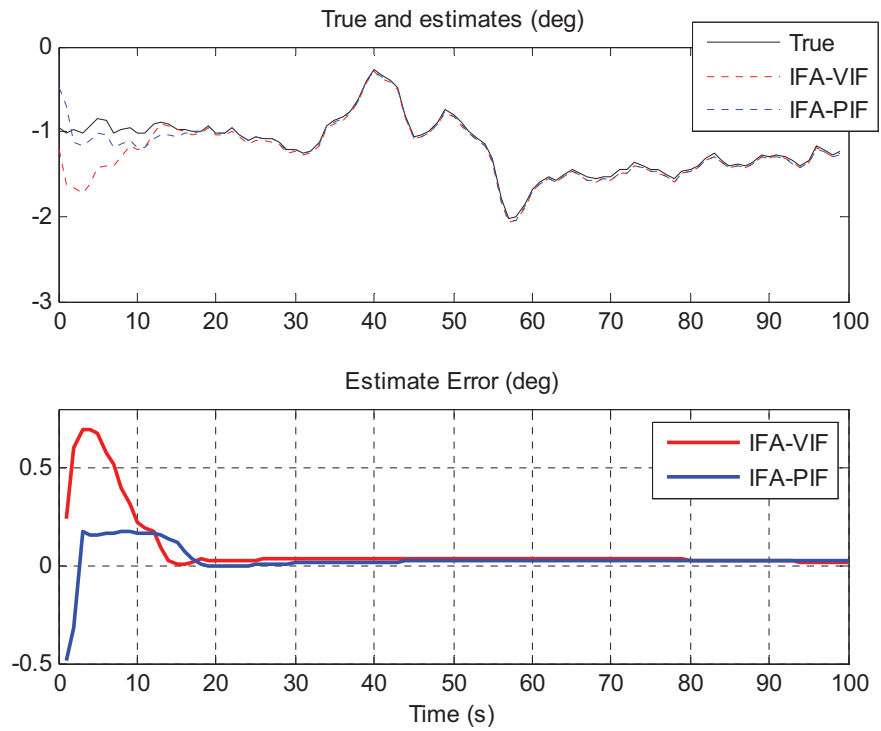

Figure 12c. Pitch angle estimate and error for S2.

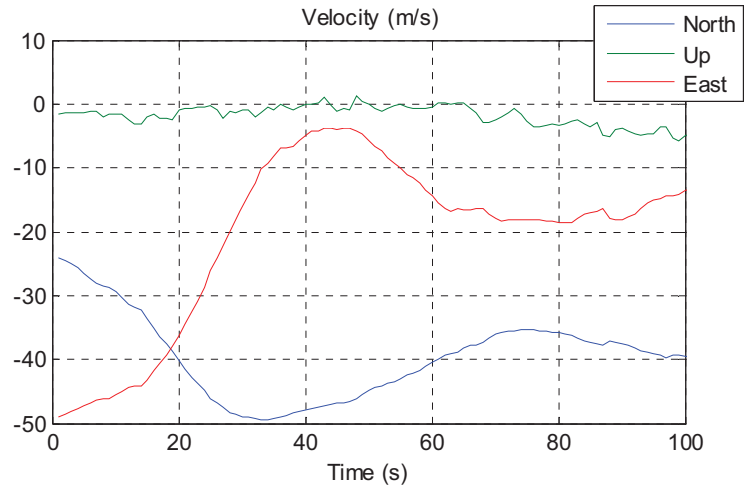

Figure 13. Velocity profile for S3.

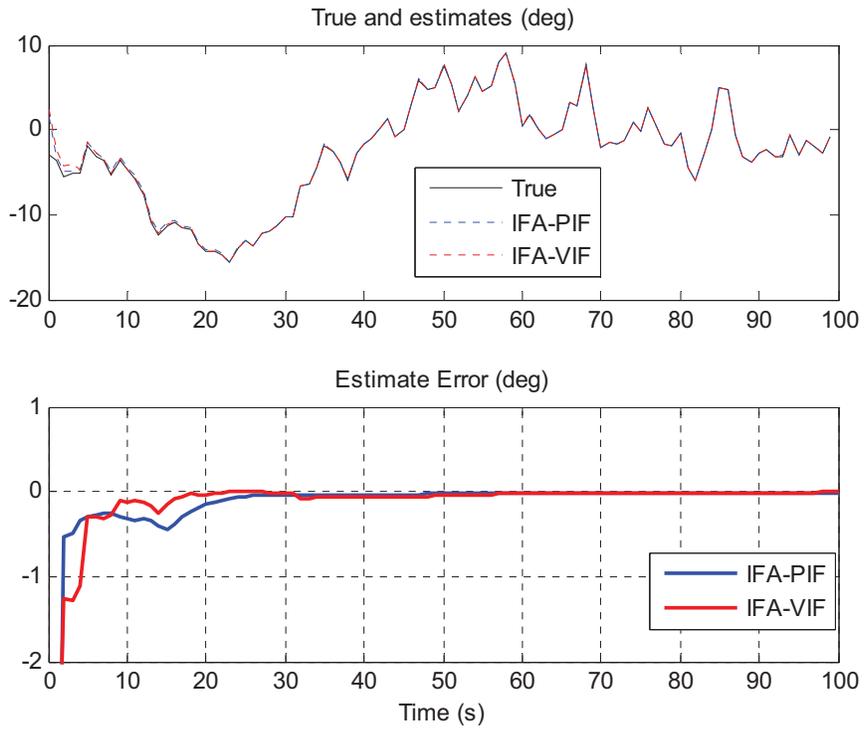

Figure 14a. Roll angle estimate and error for S3.

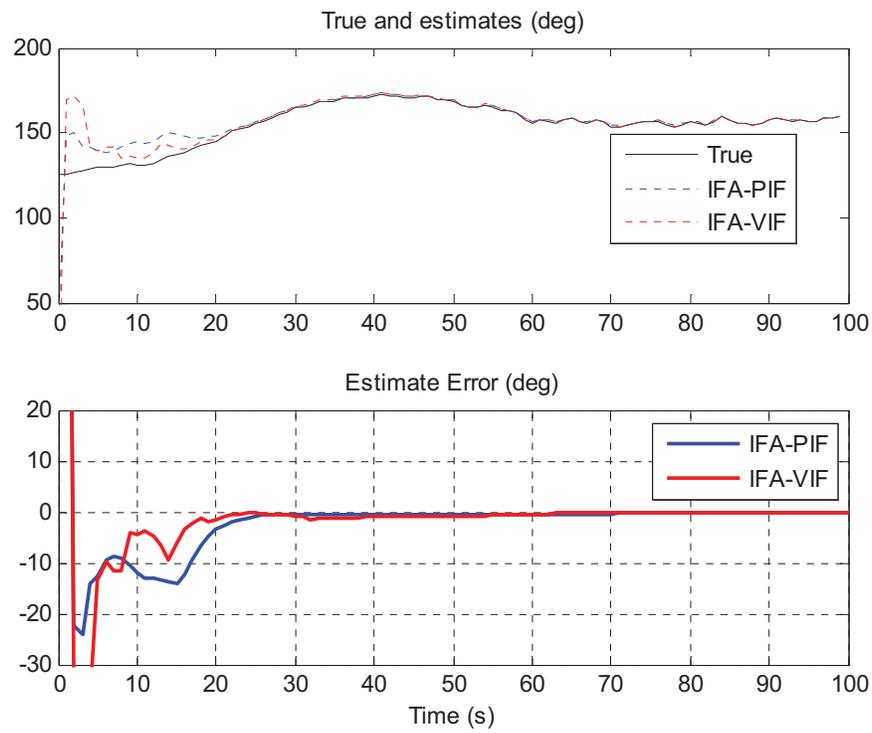

Figure 14b. Yaw angle estimate and error for S3.

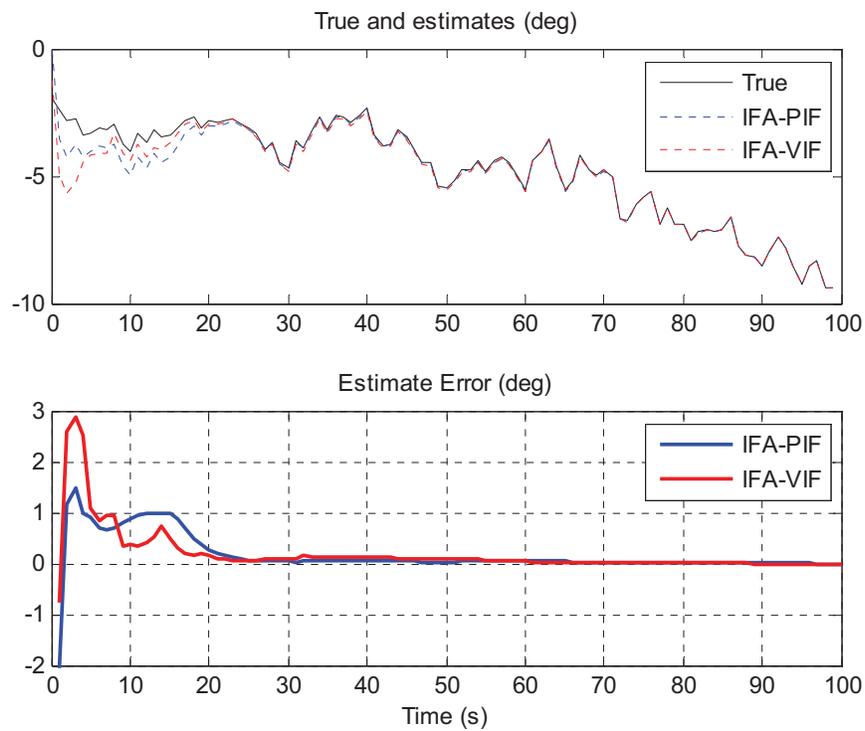

Figure 14c. Pitch angle estimate and error for S3.

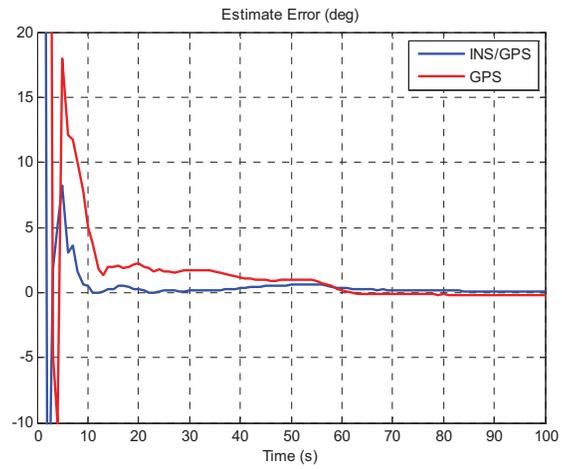

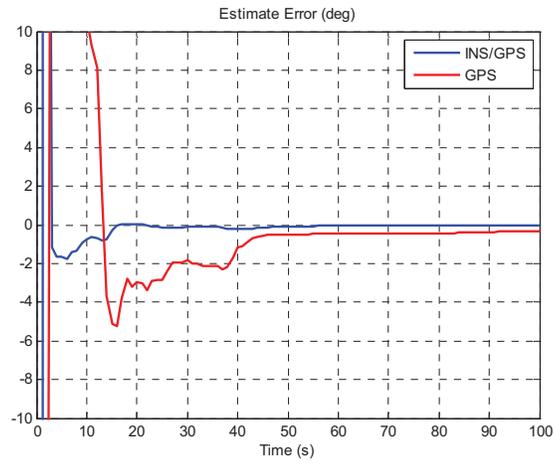

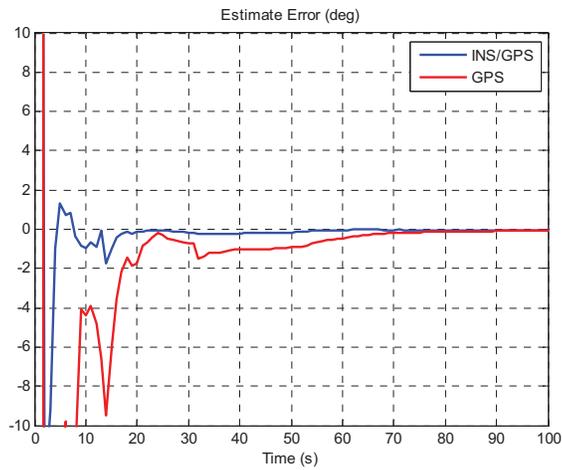

Figure 15. Yaw estimate by IFA-VIF using INS/GPS reference velocity/position:
S1 (top), S2 (middle) and S3 (bottom).

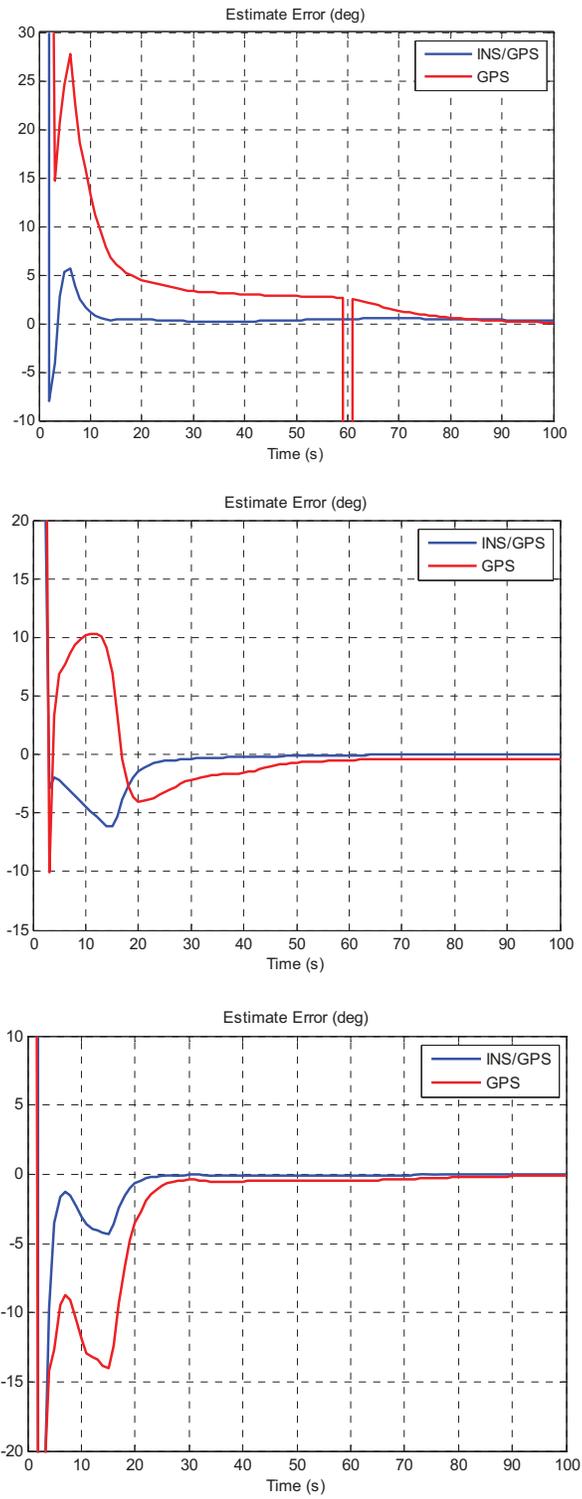

Figure 16. Yaw estimate by IFA-PIF using INS/GPS reference velocity/position:
S1 (top), S2 (middle) and S3 (bottom).

Table III-a. Roll error summary for GPS and INS/GPS

| Roll Error (deg) | | | 5s | 10s | 20s | 100s |
|---|---|---|---|---|---|---|
| IFA-VIF | GPS | S1 | 0.12 | 0.13 | 0.09 | -0.02 |
| | | S2 | 0.20 | 0.004 | -0.007 | -0.003 |
| | | S3 | -0.30 | -0.12 | -0.04 | -0.007 |
| | INS/GPS | S1 | -0.10 | -0.09 | -0.03 | 0.01 |
| | | S2 | -0.08 | -0.04 | -0.01 | 0.007 |
| | | S3 | -0.02 | 0.003 | 0.01 | 0.002 |
| IFA-PIF | GPS | S1 | 0.37 | 0.21 | 0.13 | -0.007 |
| | | S2 | 0.05 | 0.04 | 0.002 | 0.009 |
| | | S3 | -0.30 | -0.32 | -0.15 | -0.01 |
| | INS/GPS | S1 | -0.18 | -0.11 | -0.06 | 0.02 |
| | | S2 | -0.16 | -0.09 | -0.04 | 0.005 |
| | | S3 | -0.07 | -0.06 | -0.02 | 0.004 |

Table III-b. Yaw error summary for GPS and INS/GPS

| Yaw Error (deg) | | | 10s | 20s | 60s | 100s |
|---|---|---|---|---|---|---|
| IFA-VIF | GPS | S1 | 5.0 | 2.22 | 0.19 | -0.23 |
| | | S2 | 10.40 | -2.95 | -0.44 | -0.30 |
| | | S3 | -4.40 | -1.76 | -0.45 | -0.069 |
| | INS/GPS | S1 | 0.52 | 0.27 | 0.33 | 0.08 |
| | | S2 | 0.72 | 0.02 | -0.04 | -0.04 |
| | | S3 | -0.96 | -0.09 | -0.04 | -0.08 |
| IFA-PIF | GPS | S1 | 13.27 | 4.48 | -2.50 (ambiguity) | 0.09 |
| | | S2 | 10.17 | -4.00 | -0.48 | -0.41 |
| | | S3 | 11.89 | -3.56 | -0.45 | -0.10 |
| | INS/GPS | S1 | 1.17 | 0.42 | 0.46 | 0.32 |
| | | S2 | -4.44 | 1.48 | -0.05 | -0.03 |
| | | S3 | -3.08 | -0.68 | -0.13 | -0.03 |

Table III-c. Pitch error summary for GPS and INS/GPS

| Roll Error (deg) | | | 5s | 10s | 20s | 100s |
|---|---|---|---|---|---|---|
| IFA-VIF | GPS | S1 | 1.1 | 0.14 | -0.02 | 0.04 |
| | | S2 | 0.67 | 0.22 | 0.03 | 0.02 |
| | | S3 | 1.10 | 0.38 | 0.17 | -0.003 |
| | INS/GPS | S1 | 0.64 | -0.02 | -0.03 | -0.009 |
| | | S2 | -0.03 | -0.01 | -0.005 | -0.006 |
| | | S3 | -0.11 | 0.07 | 0.006 | -0.003 |
| IFA-PIF | GPS | S1 | 1.63 | 0.68 | 0.05 | 0.06 |
| | | S2 | 0.16 | 0.17 | -0.005 | 0.02 |
| | | S3 | 0.90 | 0.87 | 0.27 | -0.002 |
| | INS/GPS | S1 | 0.41 | 0.03 | -0.03 | 0.009 |
| | | S2 | -0.02 | -0.05 | -0.02 | -0.006 |
| | | S3 | 0.22 | 0.22 | 0.05 | -0.005 |